\documentclass[10pt,twocolumn,letterpaper]{article}

\usepackage{cvpr}              

\usepackage[dvipsnames]{xcolor}

\definecolor{cvprblue}{rgb}{0.21,0.49,0.74}
\usepackage[pagebackref,breaklinks,colorlinks,citecolor=cvprblue]{hyperref}
\usepackage{overpic}
\usepackage{float}
\usepackage{bm}
\usepackage{multicol}
\usepackage{multirow}
\usepackage{pifont}
\usepackage{color}
\usepackage{multirow}
\usepackage{overpic}
\usepackage{arydshln}
\usepackage{makecell}
\usepackage{times}
\usepackage{epsfig}
\usepackage[T1]{fontenc}
\usepackage{hhline}
\usepackage{xcolor}
\usepackage{pifont}
\usepackage{enumitem}
\usepackage{colortbl}
\usepackage{verbatim}
\usepackage{bm}
\usepackage{fancyhdr}
\usepackage[accsupp]{axessibility}
\definecolor{darkpink}{rgb}{0.91, 0.33, 0.5}
\definecolor{mygray}{gray}{.94}

\newcommand\blfootnote[1]{%
  \begingroup
  \renewcommand\thefootnote{}\footnote{#1}%
  \addtocounter{footnote}{-1}%
  \endgroup
}
\title{LEMON: Learning 3D Human-Object Interaction Relation from 2D Images}

\author{Yuhang Yang$^{1}$, Wei Zhai$^{1,\dagger}$, Hongchen Luo$^{1}$, Yang Cao$^{1,2}$, Zheng-Jun Zha$^{1}$\\
{$^{1}$~University of Science and Technology of China} \\
{$^{2}$~Institute of Artificial Intelligence, Hefei Comprehensive National Science Center}\\
\small{\texttt{\{yyuhang@mail., wzhai056@, lhc12@mail., forrest@, zhazj@\}ustc.edu.cn}}
}

\begin{document}

\twocolumn[{%
         \renewcommand\twocolumn[1][]{#1}%
         \maketitle
         \begin{center}
            \centering
            \vspace{-20pt}
            \includegraphics[width=0.98\textwidth]{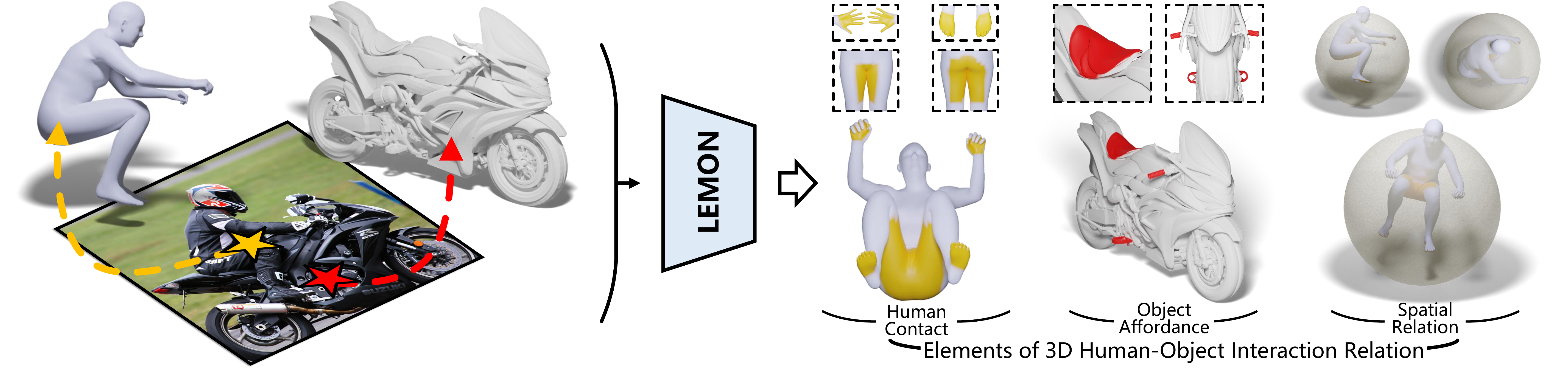}
            \captionof{figure}{For an interaction image with paired geometries of the human and object, LEMON learns 3D human-object interaction relation by jointly anticipating the interaction elements, including human contact, object affordance, and human-object spatial relation. Vertices in \textcolor[rgb]{0.88,0.82,0.36}{yellow} denote those in contact with the object, regions in \textcolor[rgb]{1,0.25,0.25}{red} are object affordance regions, and the \textcolor[rgb]{0.749,0.737,0.69}{translucent} sphere is the object proxy.}
            \label{fig:intro}
         \end{center}
}]
\begin{abstract}
\blfootnote{$\dagger$Corresponding Author.}
Learning 3D human-object interaction relation is pivotal to embodied AI and interaction modeling. Most existing methods approach the goal by learning to predict isolated interaction elements, \eg, human contact, object affordance, and human-object spatial relation, primarily from the perspective of either the human or the object. Which underexploit certain correlations between the interaction counterparts (human and object), and struggle to address the uncertainty in interactions. Actually, objects' functionalities potentially affect humans' interaction intentions, which reveals what the interaction is. Meanwhile, the interacting humans and objects exhibit matching geometric structures, which presents how to interact. In light of this, we propose harnessing these inherent correlations between interaction counterparts to mitigate the uncertainty and jointly anticipate the above interaction elements in 3D space. To achieve this, we present \textbf{LEMON} (\textbf{LE}arning 3D hu\textbf{M}an-\textbf{O}bject i\textbf{N}teraction relation), a unified model that mines interaction intentions of the counterparts and employs curvatures to guide the extraction of geometric correlations, combining them to anticipate the interaction elements. Besides, the \textbf{3D} \textbf{I}nteraction \textbf{R}elation dataset (\textbf{3DIR}) is collected to serve as the test bed for training and evaluation. Extensive experiments demonstrate the superiority of LEMON over methods estimating each element in isolation. The code and dataset are available at \href{https://yyvhang.github.io/LEMON/}{https://yyvhang.github.io/LEMON}.
\end{abstract}
\vspace{-10pt}    
\section{Introduction}
\label{sec:intro}
Learning 3D human-object interaction (HOI) relation seeks to capture semantic co-occurrence and geometric compatibility between humans and objects in 3D space \cite{yao2010modeling, wei2013modeling, chen2019holistic++}. \emph{How can machines learn the interaction relation?} One possible solution involves perceiving certain elements capable of revealing the interaction. Contact \cite{chen2023hot, Huang:CVPR:2022, tripathi2023deco, xiao2023unified}, affordance \cite{nagarajan2020learning, deng20213d, zhao2023dualafford, geng2023gapartnet}, and spatial relation \cite{han2023chorus, li2020detailed, petrov2023popup} that elucidate ``where'' the interaction manifests between the human and object garner great attention. Capturing representations of such elements is pivotal for applications like AR/VR \cite{cheng2013affordances}, imitation learning \cite{argall2009survey, hussein2017imitation}, embodied AI \cite{savva2019habitat, sheridan2016human, grigore2013joint}, and interaction modeling \cite{hassan2021populating, ye2022scene}.

\par Humans predominantly manipulate and interact with objects in 3D space. Thus, many methods devise task-specific models to anticipate certain interaction elements, thereby perceiving 3D HOI relation. Methods \cite{Huang:CVPR:2022, shimada2022hulc, tripathi2023deco} estimate dense \textbf{human contact} based on the interaction semantics depicted in images. Some studies anticipate the \textbf{object affordance} according to objects' structures or 2D visuals \cite{deng20213d, mo2021where2act, wang2022adaafford, Yang_2023_ICCV}. Several works delve into predicting the human-object \textbf{spatial relation} through synthetic images \cite{han2023chorus} or posed human geometries \cite{petrov2023popup}. A prevailing trend in these methods involves taking certain attributes (\eg, appearances, geometries) of either humans or objects to predict an isolated interaction element, capturing one aspect of the interaction relation. However, they overlook the impact of factors mutually determined by humans and objects on interactions, such as the interaction intention and geometric correlation, which leads to struggles in addressing the uncertainty within interactions. Specifically, the uncertainty could be attributed to two principal aspects. On the one hand, the diversity of HOIs introduces challenges in capturing explicit interaction intentions solely centering on humans or objects, raising the intention uncertainty. On the other hand, the limited view and mutual occlusions give rise to invisible interaction regions in images, complicating the constitution of linkages between these regions and target 3D geometries, causing the region uncertainty. These uncertainties may culminate in ambiguous anticipations.

\par
\begin{figure}
\centering
\small
\begin{overpic}[width=0.93\linewidth]{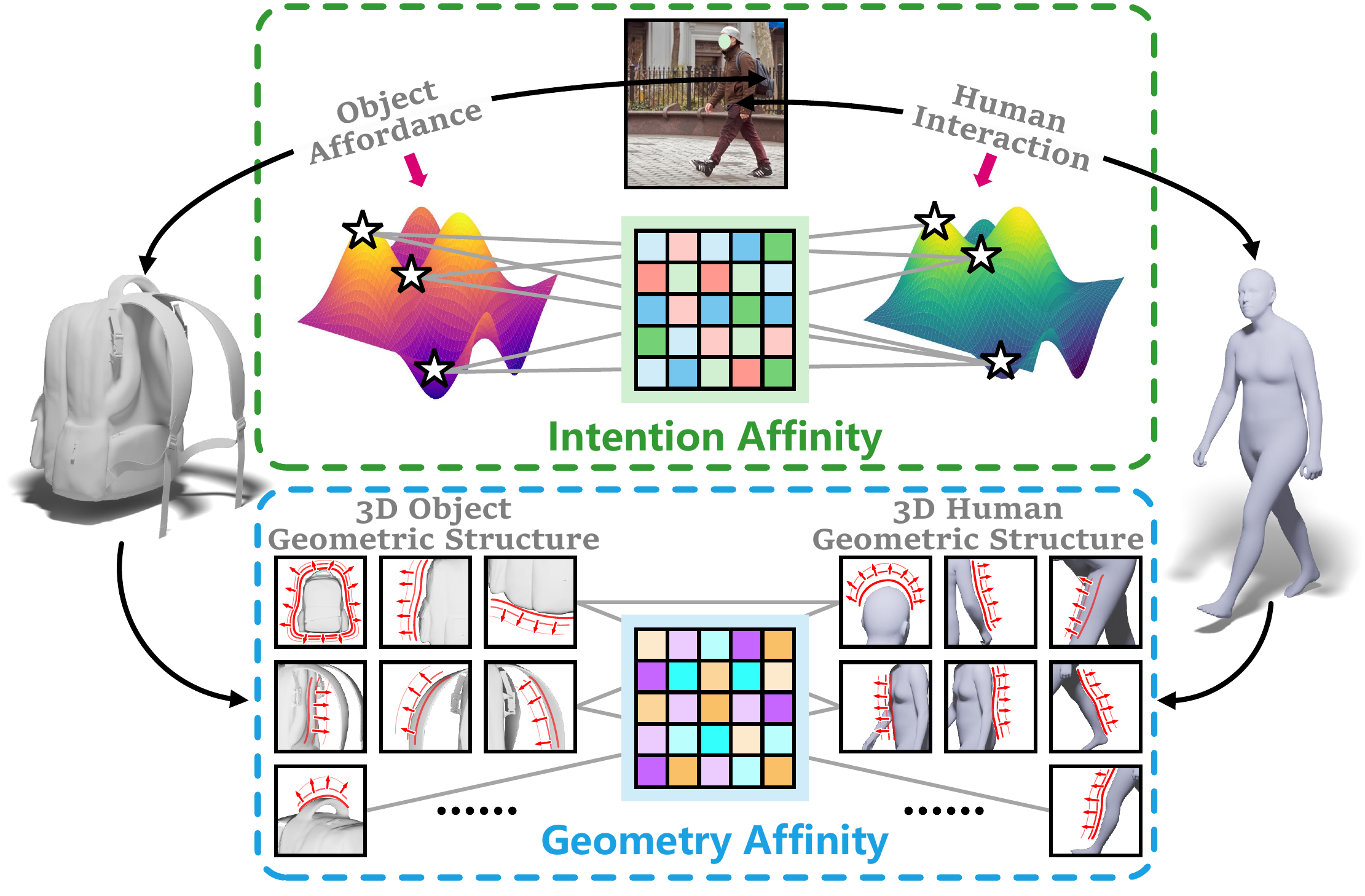}
\end{overpic}
\caption{\textbf{Motivation.} Affinities within the HOI. The object affordance inherently reveals the human's interaction intention, arising the intention affinity. The interacting human and object possess matching structures, exhibiting the geometry affinity.}
\label{fig:motivation}
\end{figure}

 In this paper, we propose leveraging both counterparts of the interaction to jointly anticipate human contact, object affordance, and human-object spatial relation in 3D space (Fig. \ref{fig:intro}), addressing the uncertainty by unearthing the correlation between the interacting humans and objects. Actually, humans and objects are intertwined and possess affinities in the interaction (Fig. \ref{fig:motivation}). In specific, the design of objects typically adheres to certain human needs. Therefore, object affordances inherently hint at ``what'' interactions humans intend to make \cite{shilling2004body, tomasello2005understanding, malle1997folk}, revealing the intention affinity of interactions. Meanwhile, the interacting human and object exhibit matching geometries (either posture or configuration), which presents ``how'' to interact, arising the geometry affinity. The intention affinity clarifies the interaction type and implicates the interaction regions. Geometry affinity could serve as pivotal clues for excavating correlations between geometries corresponding to invisible regions in the image. These interaction-related regions \eg, contact regions, further reflect the human-object spatial relation.

\par To achieve this, we present the \textbf{LEMON}, a novel framework that correlates the intention semantics and geometric correspondences to jointly anticipate human contact, object affordance, and human-object spatial relation, in 3D space. To capture the intention affinity, LEMON employs multi-branch attention to model the correlation between the interaction content in images and geometries of humans and objects, revealing intention representations of the interaction corresponding to geometries. The cosine similarity is utilized to further ensure their semantic consistency. Taking intention representations as conditions, LEMON integrates geometric curvatures to capture the geometry affinity and reveal the human contact and object affordance representations. These representations then assist in anticipating the spatial relation constrained by a combined distance loss.

\par In addition, we collect the \textbf{3DIR} dataset, which contains natural HOI images paired with object point clouds and SMPL-H \cite{MANO:SIGGRAPHASIA:2017} pseudo-GTs. Multiple annotations are made for these data, \eg, dense human contact, object affordance, and human-object spatial relation. It serves as the test bed for the model training and evaluation. 

\par The contributions are summarized as follows:
\begin{itemize}[leftmargin=15pt,topsep=0pt,itemsep=2pt]
    \item[\textbf{1)}] We thoroughly exploit the correlation between the interaction counterparts to jointly anticipate human contact, object affordance, and human-object spatial relation in 3D space. It furnishes essential interaction elements to comprehend 3D HOIs.
    \item[\textbf{2)}] We present the LEMON, a novel framework that correlates semantic interaction intention and geometries to capture the affinity between humans and objects, eliminating the impact of the interaction uncertainty and anticipating plausible 3D interaction elements.
    \item[\textbf{3)}] We provide the 3DIR dataset that contains paired HOI data and multiple annotations, including dense human contact, object affordance, and human-object spatial relation, to support the anticipation. Extensive experiments demonstrate the superiority of LEMON.
\end{itemize}

\section{Related Work}
\label{sec:related work}

\subsection{Human-Object Interaction Relation in 2D}
Various approaches perceive elements pertinent to human-object interaction relation from multiple perspectives in the 2D domain. Methods \cite{xu2019learning, zou2021end, chao2018learning, gkioxari2018detecting, li2023neural} represent human-object interaction as triplets $<$human, action, object$>$. Grounding bounding boxes of interaction counterparts and actions corresponding to them. They output the instance-level human-object spatial relation in pixel space and action semantics. Delving into the object side, the concept of ``interaction possibility'' is encapsulated in the term affordance \cite{gibson2014ecological}, and bountiful methods explore to anticipate the object affordance, whether at instance-level or part-level \cite{do2018affordancenet, nagarajan2019grounded, fang2018demo2vec, Luo_2022_CVPR, Luo_2023_CVPR, zhai2022one}. Turning to the human side, the contact explicitly indicates where the human interacts with objects. Some works detect the contact for specific parts (\eg, hand and foot) at a box-region level \cite{shan2020understanding, narasimhaswamy2020detecting, rempe2020contact}. HOT \cite{chen2023hot} provides 2D contact annotations to support human-object contact estimation in images. These efforts achieve substantial results in understanding the human-object interaction relation in 2D space. However, they still encounter challenges when extrapolating to 3D space for practical applications due to the lack of a dimension.

\subsection{Human-Object Interaction Relation in 3D}
To facilitate the incorporation of interaction comprehension into applications, extensive works explore to anticipate interaction elements in 3D space. For object affordance, methods \cite{deng20213d, xu2022partafford, wang2022adaafford, mo2021where2act, wu2023learning, zhai2023background, zhao2022dualafford} constitute the linkage between object shapes and affordances by a learning-based mapping, approaching the matter from the perspective of objects' functionality and structure. Some approaches learn object affordances from a distinct perspective, based on interactions. Whether making agents actively interact with 3D synthetic scenarios \cite{nagarajan2020learning}, or ground 3D object affordances through object-object \cite{mo2022o2o} and 2D \cite{Yang_2023_ICCV} interactions. 

\par The dense human contact estimation is primarily based on the SMPL series \cite{loper2023smpl, SMPL-X:2019, MANO:SIGGRAPHASIA:2017}. Numerous methods model human contact for various tasks \cite{clever2020bodies, huang2023diffusion, xie2022chore, yi2023mime, yin2023rotating, hassan2021populating, huang2022intercap}. HULC \cite{shimada2022hulc}, BSTRO \cite{Huang:CVPR:2022}, and DECO \cite{tripathi2023deco} are more similar to the estimation of contact in our method. They learn a mapping from the interaction semantics in images to the human vertices sequence, hardly leveraging geometries. Different from them, our method harnesses semantic and geometric correlations of interaction counterparts to address the uncertainty and infer the vertices that contact with objects. DECO also contributes the DAMON \cite{tripathi2023deco} dataset that possesses dense 3D contact annotation for in-the-wild images. In addition to human contact annotations, 3DIR includes other annotations related to the interaction, \eg, 3D object affordance and human-object spatial relation. 

\par Bounding boxes obtained by HOI detection methods give the human-object spatial relation in pixel space. To lift this relation to 3D space, DJ-RN \cite{li2020detailed} takes hollow spheres with defined radii to represent objects, and projects the spatial relation in images to 3D space based on the bounding boxes and defined radii. CHORUS \cite{han2023chorus} utilizes synthetic multi-view images generated by the diffusion model \cite{rombach2022high} to learn the spatial relation between objects and canonical humans. Given a posed human, Object Pop-Up \cite{petrov2023object} anticipates the objects and their spatial positions to match the human body for certain interactions.

\par The above methods commonly focus on one side of the interaction in isolation. In contrast, our method captures the inherent affinity between both sides of the interaction to jointly anticipate interaction elements in 3D space. Such anticipations are beneficial for tasks \eg, robot manipulation \cite{bicchi2000robotic, mandikal2021learning, zhu2023diff}, interaction generation \cite{kulkarni2023nifty, ye2022scene, li2023object, xu2023interdiff, Zhao:ICCV:2023} and reconstruction \cite{xie2022chore, xu2021d3d, hassan2021populating, zhang2020perceiving}.

\section{Method}
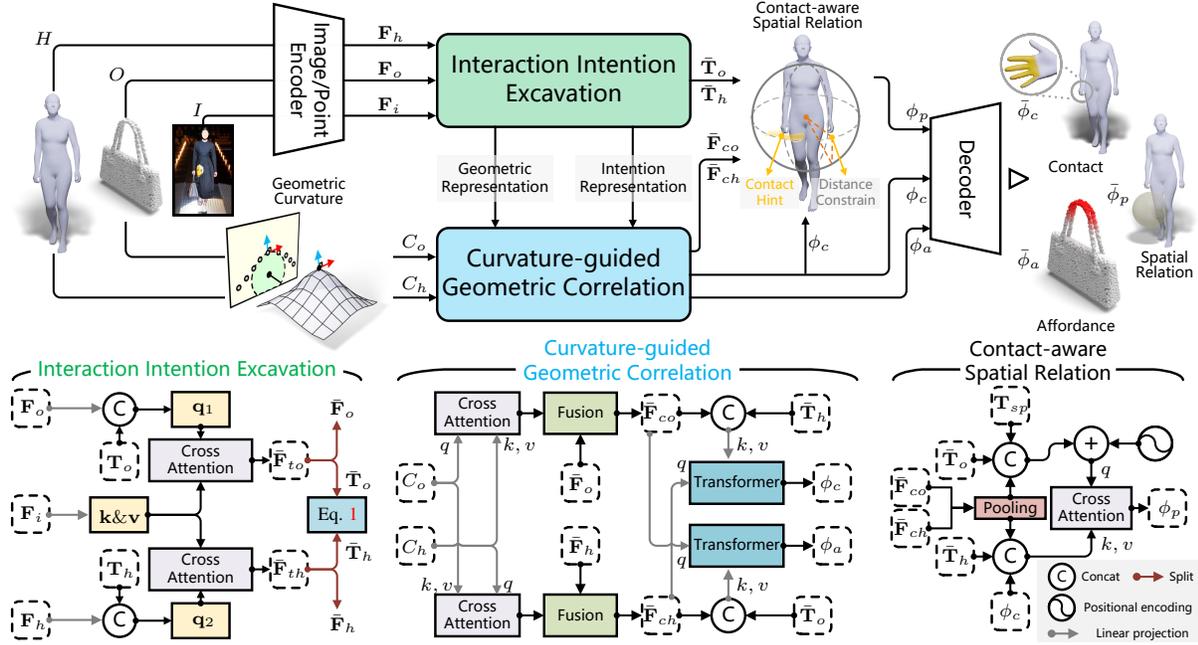
\begin{figure*}[t]
	\centering
        \scriptsize
	\begin{overpic}[width=0.91\linewidth]{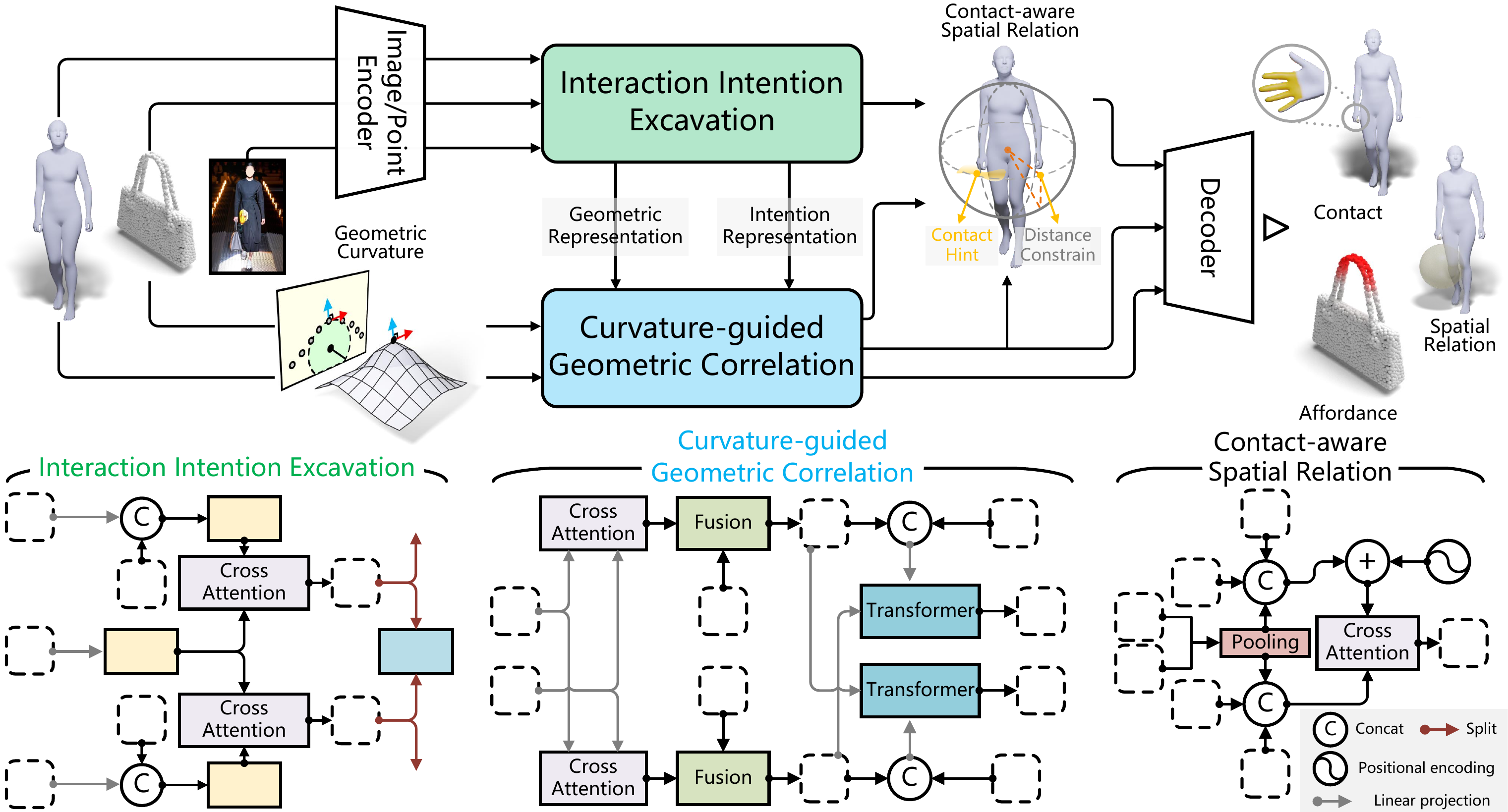}
 
        \put(2.3,50){\scriptsize$H$}
        \put(8.5,47){\scriptsize$O$}
        \put(15.5,44){\scriptsize$I$}

        \put(31,50.5){\scriptsize$\mathbf{F}_{h}$}
        \put(31,47.5){\scriptsize$\mathbf{F}_{o}$}
        \put(31,44.5){\scriptsize$\mathbf{F}_{i}$}

        \put(58,47.6){\scriptsize$\bar{\mathbf{T}}_{o}$}
        \put(58,45.3){\scriptsize$\bar{\mathbf{T}}_{h}$}

        \put(58.5,41.3){\scriptsize$\bar{\mathbf{F}}_{co}$}
        \put(58.5,38.7){\scriptsize$\bar{\mathbf{F}}_{ch}$}

        \put(33,33){\scriptsize$C_{o}$}
        \put(33,29.5){\scriptsize$C_{h}$}
 
        \put(1.1,19.2){\scriptsize$\mathbf{F}_{o}$}
        \put(1.1,10.3){\scriptsize$\mathbf{F}_{i}$}
        \put(1.1,1.3){\scriptsize$\mathbf{F}_{h}$}

        \put(8.3,14.5){\scriptsize$\mathbf{T}_{o}$}
        \put(8.3,5.6){\scriptsize$\mathbf{T}_{h}$}
        \put(7.7,10){\scriptsize$\mathbf{k \& v}$}
        
        \put(15.5,19.3){\scriptsize$\mathbf{q}_{1}$}
        \put(15.5,1.5){\scriptsize$\mathbf{q}_{2}$}

        \put(22.2,14.7){\scriptsize$\bar{\mathbf{F}}_{to}$}
        \put(22.2,5.6){\scriptsize$\bar{\mathbf{F}}_{th}$}
        
        \put(26,10.2){\scriptsize Eq. \ref{Eq:1}}
        
        \put(27,19){\scriptsize$\bar{\mathbf{F}}_{o}$}
        \put(27,1){\scriptsize$\bar{\mathbf{F}}_{h}$}
        \put(28.3,13){\scriptsize$\bar{\mathbf{T}}_{o}$}
        \put(28.3,7){\scriptsize$\bar{\mathbf{T}}_{h}$}

        \put(33,13){\scriptsize$C_{o}$}
        \put(33,7.5){\scriptsize$C_{h}$}
        \put(36.2,16){\scriptsize$q$}
        \put(41.5,16){\scriptsize$k,v$}
        \put(34.6,4.3){\scriptsize$k,v$}
        \put(41.5,4.5){\scriptsize$q$}

        \put(47,12.7){\scriptsize$\bar{\mathbf{F}}_{o}$}
        \put(47,7.5){\scriptsize$\bar{\mathbf{F}}_{h}$}

        \put(53.2,18.7){\scriptsize$\bar{\mathbf{F}}_{co}$}
        \put(53.2,1.7){\scriptsize$\bar{\mathbf{F}}_{ch}$}
        \put(56,14.5){\scriptsize$q$}
        \put(56,6.5){\scriptsize$q$}
        \put(61,16){\scriptsize$k,v$}
        \put(61,4.3){\scriptsize$k,v$}

        \put(66.3,18.7){\scriptsize$\bar{\mathbf{T}}_{h}$}
        \put(66.3,1.7){\scriptsize$\bar{\mathbf{T}}_{o}$}

        \put(68,12.9){\scriptsize$\phi_{c}$}
        \put(68,7.7){\scriptsize$\phi_{a}$}

        \put(74.2,12.5){\scriptsize$\bar{\mathbf{F}}_{co}$}
        \put(74.2,9){\scriptsize$\bar{\mathbf{F}}_{ch}$}
        \put(78.2,14.8){\scriptsize$\bar{\mathbf{T}}_{o}$}
        \put(78.2,6.7){\scriptsize$\bar{\mathbf{T}}_{h}$}
        \put(83,2.1){\scriptsize$\phi_{c}$}
        \put(82.5,19.5){\scriptsize$\mathbf{T}_{sp}$}
        \put(96.3,10.8){\scriptsize$\phi_{p}$}
        \put(91.5,14){\scriptsize$q$}
        \put(91.5,7.7){\scriptsize$k,v$}

        \put(67,33){\scriptsize$\phi_{c}$}
        \put(75.5,32.8){\scriptsize$\phi_{a}$}
        \put(75.2,37){\scriptsize$\phi_{c}$}
        \put(75.2,44){\scriptsize$\phi_{p}$}

        \put(84.5,44){\scriptsize$\bar{\phi}_{c}$}
        \put(92,37){\scriptsize$\bar{\phi}_{p}$}
        \put(84.5,31.5){\scriptsize$\bar{\phi}_{a}$}
	\end{overpic}
	\caption{\textbf{LEMON pipeline.} Initially, it takes modality-specific backbones to extract respective features $\mathbf{F}_{h}, \mathbf{F}_{o}, \mathbf{F}_{i}$, which are then utilized to excavate intention features ($\bar{\mathbf{T}}_{o}, \bar{\mathbf{T}}_{h}$) of the interaction (Sec. \ref{Sec.3.2}). With $\bar{\mathbf{T}}_{o}, \bar{\mathbf{T}}_{h}$ as conditions, LEMON integrates curvatures ($C_o, C_h$) to model geometric correlations and reveal the contact $\phi_{c}$, affordance $\phi_{a}$ features (Sec. \ref{Sec.3.3}). Following, the $\phi_{c}$ is injected into the calculation of the object spatial feature $\phi_{p}$ (Sec. \ref{Sec.3.4}). Eventually, the decoder projects $\phi_{c}, \phi_{a}, \phi_{p}$ to the final outputs $\bar{\phi}_{c}, \bar{\phi}_{a}, \bar{\phi}_{p}$.}
 \label{fig:method}
\end{figure*}

Given the inputs $\{H,O,I\}$, where $H \in \mathbb{R}^{N_{h} \times 3}$ indicates vertices of the SMPL-H \cite{MANO:SIGGRAPHASIA:2017}, which represents the human as pose $\theta \in \mathbb{R}^{52 \times 3}$, shape $\beta \in \mathbb{R}^{10}$ parameters, and output a mesh $M(\theta, \beta) \in \mathbb{R}^{6890\times3}$. $O \in \mathbb{R}^{N_{o} \times 3}$ is an object point cloud, $I \in \mathbb{R}^{H \times W \times 3}$ is an image. $N_{h}, N_{o}$ are the number of points, $H,W$ are image's height and width. LEMON jointly anticipates the human contact $\bar{\phi}_{c} \in \mathbb{R}^{N_{h} \times 1}$, object affordance $\bar{\phi}_{a} \in \mathbb{R}^{N_{o} \times 1}$, and object center position $\bar{\phi}_{p} \in \mathbb{R}^{3}$, in 3D space. As shown in Fig. \ref{fig:method}, initially, the inputs are sent to image and point cloud backbones, obtain respective features $\mathbf{F}_{h}, \mathbf{F}_{o}, \mathbf{F}_{i}$. Then, LEMON correlates $\mathbf{F}_{i}$ with $\mathbf{F}_{o}$, $\mathbf{F}_{h}$ to extract features of the interaction intention ($\bar{\mathbf{T}}_{o}, \bar{\mathbf{T}}_{h}$) inherent in geometries through multi-branch attention, and employs cosine similarity to constrain their semantic consistency (Sec. \ref{Sec.3.2}). With $\bar{\mathbf{T}}_{o}$, $\bar{\mathbf{T}}_{h}$ as conditions, LEMON integrates curvatures to guide the modeling of geometric correlations, capturing the geometry affinity and revealing the contact $\phi_{c}$ and affordance $\phi_{a}$ features (Sec. \ref{Sec.3.3}). Next, geometric features and $\phi_{c}, \bar{\mathbf{T}}_{o}, \bar{\mathbf{T}}_{h}$ are utilized to model the object spatial feature $\phi_{p}$ (Sec. \ref{Sec.3.4}). Eventually, $\phi_{c}, \phi_{a}, \phi_{p}$ are projected to $\bar{\phi}_{c}$, $\bar{\phi}_{a}$, and $\bar{\phi}_{p}$ in the decoder, the whole process is optimized by a combined loss (Sec. \ref{Sec.3.5}).

\subsection{Interaction Intention Excavation}
\label{Sec.3.2}
We add human and object masks on $I$, and utilize the HRNet \cite{wang2020deep} and DGCNN \cite{wang2019dynamic} as backbones to extract the image feature $\mathbf{F}_{i} \in \mathbb{R}^{C \times h \times w}$, the geometric features of human $\mathbf{F}_{h} \in \mathbb{R}^{C \times N_{h}}$, and object $\mathbf{F}_{o} \in \mathbb{R}^{C \times N_{o}}$, $\mathbf{F}_{i}$ is flattened to $\mathbb{R}^{C \times hw}$. Images contain rich interaction semantics, which could serve as clues to unearth the interaction intention within the geometries. In detail, tokens $\mathbf{T}_{o}, \mathbf{T}_{h} \in \mathbb{R}^{C \times 1}$ are generated to represent intention features inherent in geometries. $\mathbf{T}_{o}, \mathbf{T}_{h}$ are concatenated with $\mathbf{F}_{o}$ and $\mathbf{F}_{h}$ respectively to get the feature sequences $\mathbf{F}_{to} \in \mathbb{R}^{C \times (N_{o}+1)}, \mathbf{F}_{th} \in \mathbb{R}^{C \times (N_{h}+1)}$. Taking $\mathbf{F}_{i}$ as the shared key and value, $\mathbf{F}_{to}$, $\mathbf{F}_{th}$ as queries ($q_{1}, q_{2}$) in two branches, the multi-branch attention is employed to model the intention features, expressed as $\bar{\mathbf{F}}_{to}, \bar{\mathbf{F}}_{th} = f_\delta([\mathbf{F}_{to}, \mathbf{F}_{th}], \mathbf{F}_{i})$. Where $f_{\delta}$ indicates the multi-branch attention layer \cite{qiu2023mb, yu2022dual, li2023g2l}, $[\cdot]$ denotes two branches with different queries. Human and object geometries possess multiple interaction possibilities, which may introduce semantic ambiguity. To mitigate this, we further constrain the consistency of semantic tokens:
\begin{equation}
\small
\label{Eq:1}
    \varphi = \frac{\bar{\mathbf{T}}_{o} \cdot \bar{\mathbf{T}}_{h}}{||\bar{\mathbf{T}}_{o}||_{2} \times ||\bar{\mathbf{T}}_{h}||_{2}},
\end{equation}
where $\bar{\mathbf{T}}_{o}, \bar{\mathbf{T}}_{h} \in \mathbb{R}^{C \times 1}$ are split from $\bar{\mathbf{F}}_{to}$, $\bar{\mathbf{F}}_{th}$, representing the semantic intention features of geometries. $\varphi$ is the semantic cosine similarity, a part of $\mathcal{L}_{s}$ in Sec. \ref{Sec.3.5}.

\subsection{Curvature-guided Geometric Correlation}
\label{Sec.3.3}
The interacting humans and objects exhibit certain geometry affinity \cite{yang2021cpf}, manifesting in matching geometric structures and correlative curvatures. For geometric curvatures, the normal curvature could better represent local structures like interaction regions \cite{nurunnabi2015outlier, koenderink1992surface}. Thus, we encode the normal curvatures into geometric features, and regard the $\bar{\mathbf{T}}_{o}, \bar{\mathbf{T}}_{h}$ as conditions to capture the affinity among human and object geometries. Normal curvatures $C_o \in \mathbb{R}^{1 \times N_{o}}$, $C_h \in \mathbb{R}^{1 \times N_{h}}$ of object and human are obtained by local fitting method \cite{zhang2009robust, xu2023globally}. To correlate the curvatures, $C_o, C_h$ are encoded to high dimension $C^{'}_{o} \in \mathbb{R}^{C \times N_{o}}, C^{'}_{h} \in \mathbb{R}^{C \times N_{h}}$, and the cross-attention $f_m$ is mutually performed on them, formulated as $\bar{C}_{o} = f_m(C^{'}_{o}, C^{'}_{h}), \bar{C}_{h} = f_m(C^{'}_{h}, C^{'}_{o})$. Following, to make the curvature guide the calculation of geometric correlation, $\bar{C}_{o}, \bar{C}_{h}$ are integrated and fused with geometric features:
\begin{equation}
\small
\bar{\mathbf{F}}_{co} = f(\Gamma(\bar{C}_{o}, \bar{\mathbf{F}}_{o})), \bar{\mathbf{F}}_{ch} = f(\Gamma(\bar{C}_{h}, \bar{\mathbf{F}}_{h})),
\end{equation}
where $\bar{\mathbf{F}}_{co},\bar{\mathbf{F}}_{o} \in \mathbb{R}^{C \times N_{o}}$ and $\bar{\mathbf{F}}_{ch},\bar{\mathbf{F}}_{h} \in \mathbb{R}^{C \times N_{h}}$. $\bar{\mathbf{F}}_{o},\bar{\mathbf{F}}_{h}$ are split from $\bar{\mathbf{F}}_{to}$ and $\bar{\mathbf{F}}_{th}$, $\Gamma$ denotes the concatenation, $f$ indicates convolution layers with $1 \times 1$ kernel. Then, $\bar{\mathbf{T}}_{o}$ and $\bar{\mathbf{T}}_{h}$ are considered as conditions \cite{meng2021conditional} to further screen candidate regions that match the interaction depicted in images and participate in the modeling of geometry affinity, revealing the affordance and contact features $\phi_{a}, \phi_{c}$. The process could be formulated as:
\begin{equation}
\small
\label{Eq:3}
\begin{split}
    \phi_{a} = f_\theta(\bar{\mathbf{F}}_{co}, \Gamma(\bar{\mathbf{F}}_{ch}, \bar{\mathbf{T}}_{o})), \phi_{c} = f_\theta(\bar{\mathbf{F}}_{ch}, \Gamma(\bar{\mathbf{F}}_{co}, \bar{\mathbf{T}}_{h})),
\end{split}
\end{equation}
where $\phi_{a} \in \mathbb{R}^{C \times N_{o}}$, to calculate it, $\bar{\mathbf{F}}_{co}$ serves as the query, and the concatenation of $\bar{\mathbf{F}}_{ch}$ and $\mathbf{\bar{T}}_{o}$ serves as the key and value, $f_{\theta}$ is the cross-transformer layer. $\phi_{c} \in \mathbb{R}^{C \times N_{h}}$ is obtained through the analogous way.

\begin{figure*}[t]
	\centering
	\small
        \begin{overpic}[width=0.94\linewidth]{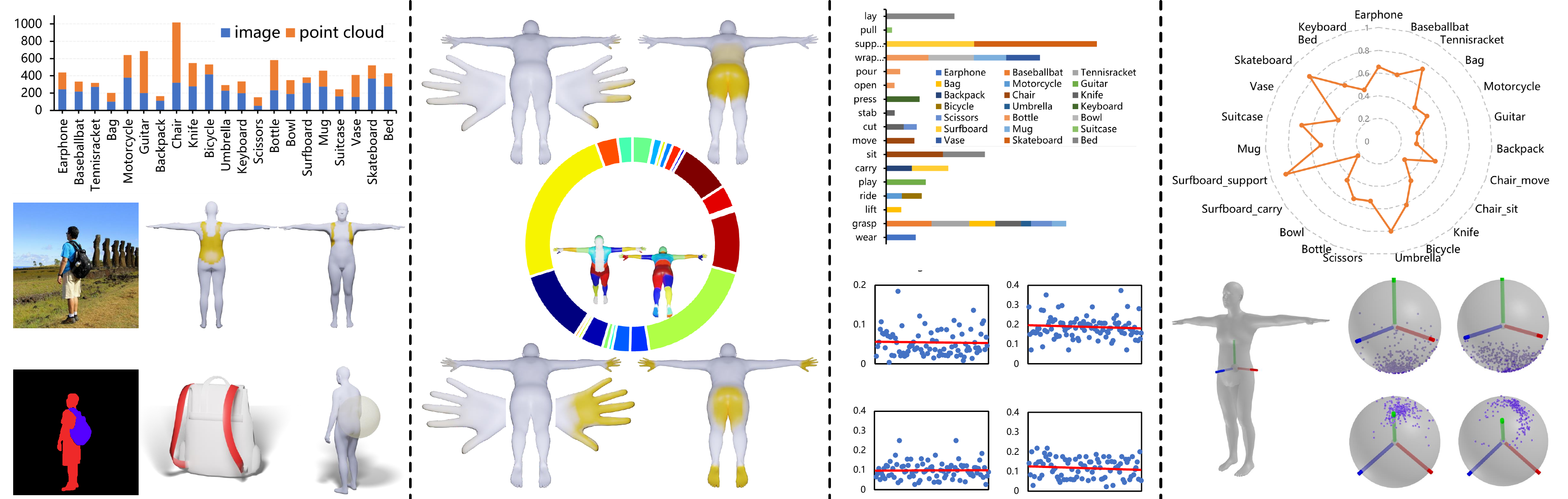}
        \put(13,-1.5){{\textbf{\footnotesize (a)}}}
        \put(39,-1.5){{\textbf{\footnotesize (b)}}}
        \put(63,-1.5){{\textbf{\footnotesize (c)}}}
        \put(84,-1.5){{\textbf{\footnotesize (d)}}}
        
        \put(3.7,9.7){{\textbf{\tiny Image}}}
        \put(3.7,-1){{\textbf{\tiny Mask}}}
        \put(11.3,9){{\textbf{\tiny Affordance}}}
        \put(16,12){{\textbf{\tiny Contact}}}
        \put(18.5,9){{\textbf{\tiny Spatial Relation}}}

        \put(28.5,22){{\textbf{\tiny Mug}}}
        \put(28.3,21){{\textbf{\tiny Grasp}}}
        \put(49,22){{\textbf{\tiny Chair}}}
        \put(49.7,21){{\textbf{\tiny Sit}}}
        \put(27.5,11){{\textbf{\tiny Tennisracket}}}
        \put(29,10){{\textbf{\tiny Grasp}}}
        \put(47.5,11){{\textbf{\tiny Motorcycle}}}
        \put(49,10){{\textbf{\tiny Ride}}}
        \put(36.5,18.5){{\textbf{\tiny Ratio of Contact}}}
        \put(38,17.5){{\textbf{\tiny on 24 Parts}}}

        \put(60,19){{\textbf{\tiny Ratio of Affordance Region}}}
        \put(57.7,14.5){{\textbf{\tiny Bag-lift}}}
        \put(65.3,14.5){{\textbf{\tiny Baseballbat-grasp}}}
        \put(56.8,6.3){{\textbf{\tiny Scissors-cut}}}
        \put(66,6.3){{\textbf{\tiny Scissors-grasp}}}
        
        \put(93.5,7){{\textbf{\tiny Motorcycle}}}
        \put(86.5,7){{\textbf{\tiny Skateboard}}}
        \put(86.9,-0.5){{\textbf{\tiny Earphone}}}
        \put(93.8,-0.5){{\textbf{\tiny Umbrella}}}
        \put(80.5,7.8){{\textbf{\scriptsize x}}}
        \put(78.5,10.3){{\textbf{\scriptsize y}}}
        \put(76.5,7.8){{\textbf{\scriptsize z}}}
        \put(76.5,0.7){{\textbf{\tiny Reference}}}
        \put(77.2,-0.2){{\textbf{\tiny Frame}}}

        \put(76.6,30){{\textbf{\tiny Distance}}}
        \put(76,29){{\textbf{\tiny Distribution}}}
        
        \put(76.5,15.5){{\textbf{\tiny Directional}}}
        \put(76.7,14.5){{\textbf{\tiny Projection}}}
	\end{overpic}
	\caption{\textbf{3DIR Dataset.} \textbf{(a)} The quantity of images and point clouds for each object, and a data sample containing the image, mask, dense human contact annotation, 3D object with affordance annotation, and the fitted human mesh with the object proxy sphere. \textbf{(b)} The proportion of our contact annotations within 24 parts on SMPL \cite{loper2023smpl}, and distributions of contact vertices for certain HOIs. \textbf{(c)} The ratio of annotated affordance regions to the whole object geometries, and the distribution of this ratio for some categories. \textbf{(d)} Mean distances (unit: m) between annotated object centers and human pelvis joints, and directional projections of annotated centers for several objects.}
 \label{Fig:dataset}
\end{figure*}
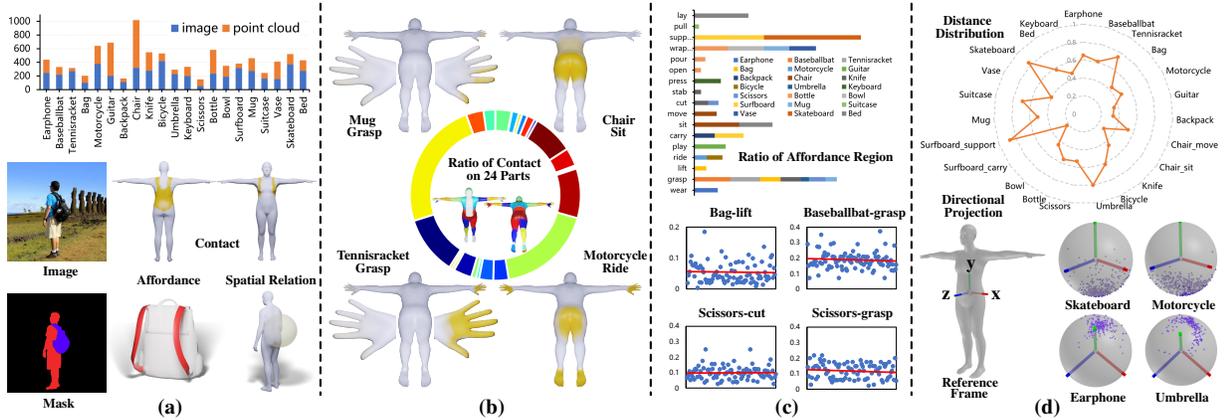

\subsection{Contact-aware Spatial Relation}
\label{Sec.3.4}
Human-object interactions are extremely diverse, rendering the reasoning of their 3D spatial relation very challenging. Nevertheless, human contact implicitly constrains the position of the object, thereby assisting in inferring plausible human-object spatial relation. Thus, the object's center is represented as a token sequence $\mathbf{T}_{sp} \in \mathbb{R}^{C \times 3}$, with introducing the contact feature as a constraint, LEMON takes the semantic intention and geometric features of the object to query the corresponding features of the human. Modeling $\mathbf{T}_{sp}$ as the object spatial feature, formulated as:
\begin{equation}
\small
    \phi_{p} = f_{\rho}(\Gamma(\Theta(\bar{\mathbf{F}}_{co}), \mathbf{\bar{T}}_{o},\mathbf{T}_{sp}) + pe, \Gamma(\Theta(\bar{\mathbf{F}}_{ch}), \mathbf{\bar{T}}_{h},\phi_{c})),
\end{equation}
where $\phi_{p} \in \mathbb{R}^{C \times 3}$, is the spatial feature of the object center, $f_{\rho}$ is a cross-attention layer, $\Theta$ denotes the pooling layer, and $pe$ indicates a learnable positional encoding.

\subsection{Loss Functions}
\label{Sec.3.5}
Eventually, $\phi_a, \phi_c, \phi_p$ are sent to the decoder that contains three projection heads with linear, normalization, and activation layers. Which output object affordance $\bar{\phi}_a \in \mathbb{R}^{N_{o} \times 1}$, human contact $\bar{\phi}_c \in \mathbb{R}^{N_{h} \times 1}$, and object's center position $\bar{\phi}_p \in \mathbb{R}^{3}$ respectively. With defined radii for specific objects, we treat a sphere as the object proxy \cite{li2020detailed}, representing the spatial relation in the camera coordinates of fitted humans. The overall training loss is expressed as:
\begin{equation}
\small
\mathbf{\mathcal{L}} = \omega_{1}\mathbf{\mathcal{L}}_{c} + \omega_{2}\mathbf{\mathcal{L}}_{a} + \omega_{3}\mathbf{\mathcal{L}}_{s} + \omega_{4}\mathbf{\mathcal{L}}_{p},
\end{equation}
where $\omega_{1-4}$ are hyper-parameters to balance the losses. $\mathbf{\mathcal{L}}_{c}$ and $\mathbf{\mathcal{L}}_{a}$ possess the same formulation, a focal loss \cite{lin2017focal} combined with a dice loss \cite{milletari2016v}. $\mathbf{\mathcal{L}}_{c}$ is calculated by $\bar{\phi}_c$ and the contact label $\hat{\phi}_c$, while $\mathbf{\mathcal{L}}_{a}$ is calculated by $\bar{\phi}_a$ and the affordance annotation $\hat{\phi}_a$. $\mathbf{\mathcal{L}}_{s}$ is employed to ensure $\bar{\mathbf{T}}_{o}$, $\bar{\mathbf{T}}_{h}$ align with interaction semantics of the image and maintain consistency in the semantic space. Specifically, $\mathbf{F}_{i}$ is mapped to a logit $y$, which is used to calculate a cross-entropy loss $\mathbf{\mathcal{L}}_{ce}$ with interaction categorical label $\hat{y}$, $\mathbf{\mathcal{L}}_{s}$ is the sum of $\mathbf{\mathcal{L}}_{ce}$ and $\varphi$ (Eq. \ref{Eq:1}). For 3D HOIs, object positions are diverse due to variations in human orientation, relying solely on absolute coordinates as supervision makes it hard for the model to learn a consistent mapping. However, the relative distances between humans and objects are similar for specific HOIs. This smoother distance distribution contributes to reducing the optimization space to an approximate sphere for specific HOIs. Consequently, we take the distance between the object center and the human pelvis joint as an additional constraint, and $\mathbf{\mathcal{L}}_{p}$ is formulated as:
\begin{equation}
\label{Eq:6}
\small
\mathbf{\mathcal{L}}_{p} = \mathbf{\mathcal{L}}_{pa} + \mathbf{\mathcal{L}}_{pr}, \quad \mathbf{\mathcal{L}}_{pa} = ||\bar{\phi}_p-\hat{\phi}_p||_{2}, \mathbf{\mathcal{L}}_{pr} = ||\bar{\xi}-\hat{\xi}||_{2},
\end{equation}
where $\hat{\phi}_p$ is the annotation of the object center. $\bar{\xi}, \hat{\xi}$ denotes distances between the pelvis joint and $\bar{\phi}_p, \hat{\phi}_p$ respectively.

\section{Dataset}

We introduce the collection and annotation protocols of the 3DIR and give some statistical analysis of the collected data and annotations, shown in Fig. \ref{Fig:dataset}.

\begin{table*}[t]
\small
\centering
  \renewcommand{\arraystretch}{1.}
  \renewcommand{\tabcolsep}{1.pt}
  \caption{{\textbf{Comparison on the 3DIR.} Evaluation metrics of comparison methods on the benchmark, the best results are covered with the mask. $\textcolor{darkpink}{\scriptstyle~\diamond}$ indicates the improvement relative to the baseline. P. means take PointNet++ \cite{qi2017pointnet++} as the backbone, and D. means DGCNN \cite{wang2019dynamic}.}}
\label{table:main_results}
\begin{tabular}{ccccc|cccc|cc}
\toprule
\multicolumn{5}{c|}{\textbf{Human Contact}}                                                                     & \multicolumn{4}{c|}{\textbf{Object Affordance}}                                            & \multicolumn{2}{c}{\textbf{Spatial Relation}}                  \\ \midrule
\multicolumn{1}{c|}{\textbf{Methods}} & \textbf{Precision $\uparrow$} & \textbf{Recall $\uparrow$} & \textbf{F1 $\uparrow$} & \textbf{geo. (cm) $\downarrow$} & \multicolumn{1}{c|}{\textbf{Methods}} & \textbf{AUC $\uparrow$} & \textbf{aIOU $\uparrow$} & \textbf{SIM $\uparrow$} & \multicolumn{1}{c|}{\textbf{Methods}} & \textbf{MSE $\downarrow$} \\ \midrule
\multicolumn{1}{c|}{Baseline}            & $0.49$              & $0.52$           & $0.49$       & $32.83$        & \multicolumn{1}{c|}{Baseline}              & $82.36$        & $32.63$         & $0.50$        & \multicolumn{1}{c|}{Baseline}            & $0.051$ \\
\multicolumn{1}{c|}{BSTRO \cite{Huang:CVPR:2022}}            & $0.57\textcolor{darkpink}{\scriptstyle~\diamond16.3\%}$              & $0.58\textcolor{darkpink}{\scriptstyle~\diamond11.5\%}$           & $0.55\textcolor{darkpink}{\scriptstyle~\diamond12.2\%}$       & $28.58\textcolor{darkpink}{\scriptstyle~\diamond12.9\%}$        & \multicolumn{1}{c|}{3DA. \cite{deng20213d}}              & $85.49\textcolor{darkpink}{\scriptstyle~\diamond3.8\%}$        & $35.42\textcolor{darkpink}{\scriptstyle~\diamond8.5\%}$         & $0.56\textcolor{darkpink}{\scriptstyle~\diamond12.0\%}$        & \multicolumn{1}{c|}{DJ-RN \cite{li2020detailed}}            & $0.042\textcolor{darkpink}{\scriptstyle~\diamond17.6\%}$       \\
\multicolumn{1}{c|}{DECO \cite{tripathi2023deco}}             & $0.70\textcolor{darkpink}{\scriptstyle~\diamond42.8\%}$           & $0.72\textcolor{darkpink}{\scriptstyle~\diamond38.4\%}$       & $0.69\textcolor{darkpink}{\scriptstyle~\diamond40.8\%}$  & $15.25\textcolor{darkpink}{\scriptstyle~\diamond53.5\%}$      & \multicolumn{1}{c|}{IAG \cite{Yang_2023_ICCV}}             & $86.63\textcolor{darkpink}{\scriptstyle~\diamond5.1\%}$      & $38.57\textcolor{darkpink}{\scriptstyle~\diamond18.2\%}$        & $0.59\textcolor{darkpink}{\scriptstyle~\diamond18.0\%}$     & \multicolumn{1}{c|}{PopUp \cite{petrov2023object}}        & $0.027\textcolor{darkpink}{\scriptstyle~\diamond47.0\%}$     \\

\midrule
\multicolumn{1}{c|}{Ours P.}            & $0.76\textcolor{darkpink}{\scriptstyle~\diamond55.1\%}$                  & $0.81\textcolor{darkpink}{\scriptstyle~\diamond55.7\%}$               & $0.77\textcolor{darkpink}{\scriptstyle~\diamond57.1\%}$           & $9.02\textcolor{darkpink}{\scriptstyle~\diamond72.5\%}$            & \multicolumn{1}{c|}{Ours P.}                & $87.91\textcolor{darkpink}{\scriptstyle~\diamond6.7\%}$            & $40.97\textcolor{darkpink}{\scriptstyle~\diamond25.5\%}$             & $0.63\textcolor{darkpink}{\scriptstyle~\diamond26.0\%}$            & \multicolumn{1}{c|}{Ours P.}                & $0.012\textcolor{darkpink}{\scriptstyle~\diamond76.4\%}$            \\

\rowcolor{mygray}
\multicolumn{1}{c|}{Ours D.}            & $0.78\textcolor{darkpink}{\scriptstyle~\diamond59.1\%}$                  & $0.82\textcolor{darkpink}{\scriptstyle~\diamond57.6\%}$               & $0.78\textcolor{darkpink}{\scriptstyle~\diamond59.1\%}$           & $7.55\textcolor{darkpink}{\scriptstyle~\diamond77.0\%}$            & \multicolumn{1}{c|}{Ours D.}                & $88.51\textcolor{darkpink}{\scriptstyle~\diamond7.4\%}$            & $41.34\textcolor{darkpink}{\scriptstyle~\diamond26.6\%}$             & $0.64\textcolor{darkpink}{\scriptstyle~\diamond28.0\%}$            & \multicolumn{1}{c|}{Ours D.}                & $0.010\textcolor{darkpink}{\scriptstyle~\diamond80.3\%}$            \\ \bottomrule
\end{tabular}
\end{table*}

\textbf{Collection.} We collect images with explicit interaction contents, in which humans interact with specific objects. These images adhere to the condition that the human's upper body is present, ensuring the efficient recovery of the human mesh. In total, we collect 5k in-the-wild images from HAKE \cite{li2020pastanet}, V-COCO \cite{gupta2015visual}, PIAD \cite{Yang_2023_ICCV}, and websites with free licenses, spanning 21 object classes and 17 interaction categories. Additionally, we collect over 5k 3D object instances from several 3D datasets \cite{mo2019partnet, deng20213d, deitke2023objaverse, liu2022akb}, based on the category of objects in collected images.

\textbf{Annotation.} We make over 25k annotations with multiple types for the collected data. \textbf{1) Masks}: With the assistance of the SAM \cite{kirillov2023segment}, we manually mask the interacting human and object in the image. \textbf{2) Human Mesh}: In natural images, some humans only exhibit the upper body. Thus, we utilize the pipeline of UBody \cite{lin2023osx} to fit SMPL-H pseudo-GTs. The 2D body and hand joints needed by the pipeline are obtained through the DWPose \cite{yang2023effective}. \textbf{3) Contact}: Similar to DAMON \cite{tripathi2023deco} and HOT \cite{chen2023hot}, the human contact is annotated based on human knowledge. For each image, we clearly provide masks and a specific interaction type for the human-object pair, professional annotators are hired to ``draw'' vertices on the human template that contact with objects. The entire annotation process cycles three rounds, with each round incorporating subjective cross-checks and objective metric checks. \textbf{4) Affordance}: We refer to the 3D-AffordanceNet \cite{deng20213d} to annotate the object affordance. The annotations of 11 objects included in 3D-AffordanceNet are directly utilized, and we annotate an additional 10 object categories excluded in 3D-AffordanceNet. \textbf{5) Spatial Relation}: For each sample, we color the fitted human mesh with annotated per-vertex human contact and treat a sphere as the object proxy. The radius of the sphere for each object category is pre-defined, referring to DJ-RN \cite{li2020detailed}. Given the posed human in color and the proxy sphere, annotators adjust the sphere relative to the human to align with the human-object spatial relation depicted in the image. Ultimately, we record the center coordinates of the adjusted sphere. Due to the page limitation, please refer to the Sup. Mat. for more annotation details.

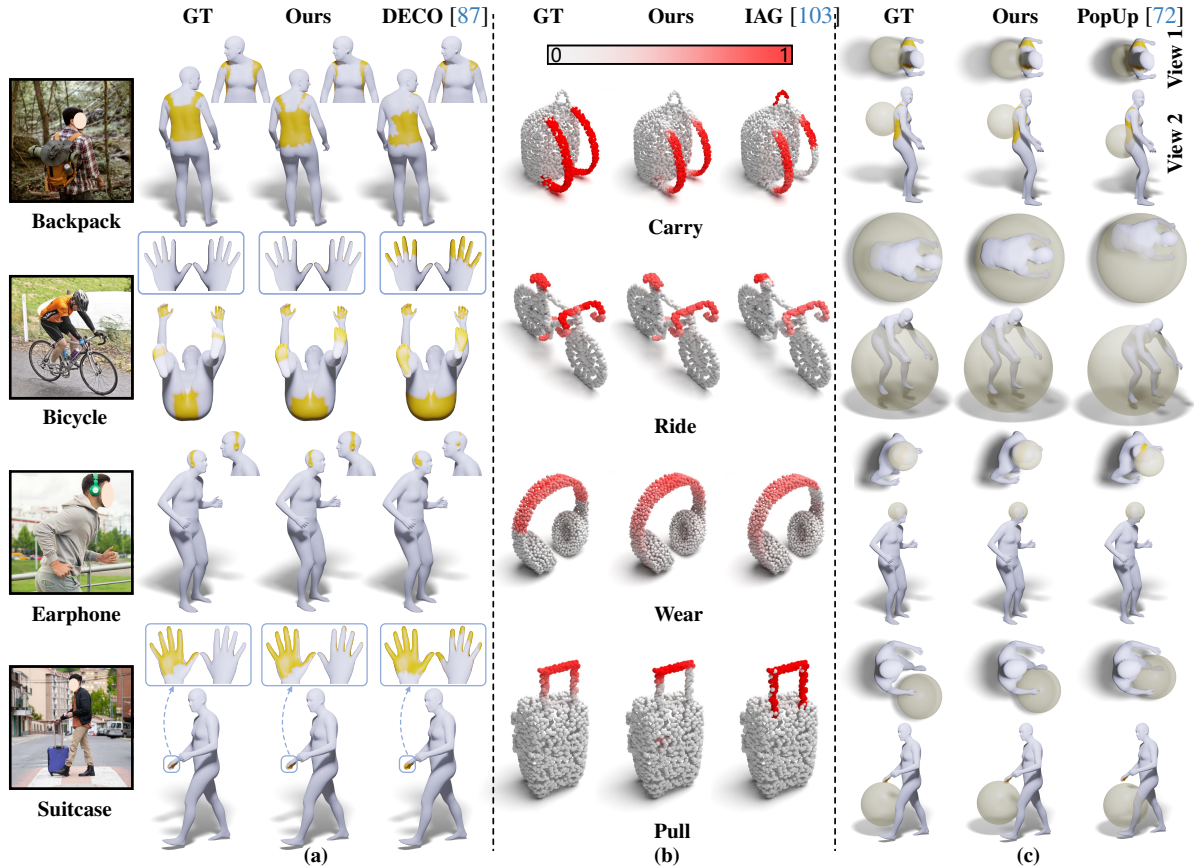
\begin{figure*}[t]
	\centering
	\small
        \begin{overpic}[width=0.92\linewidth]{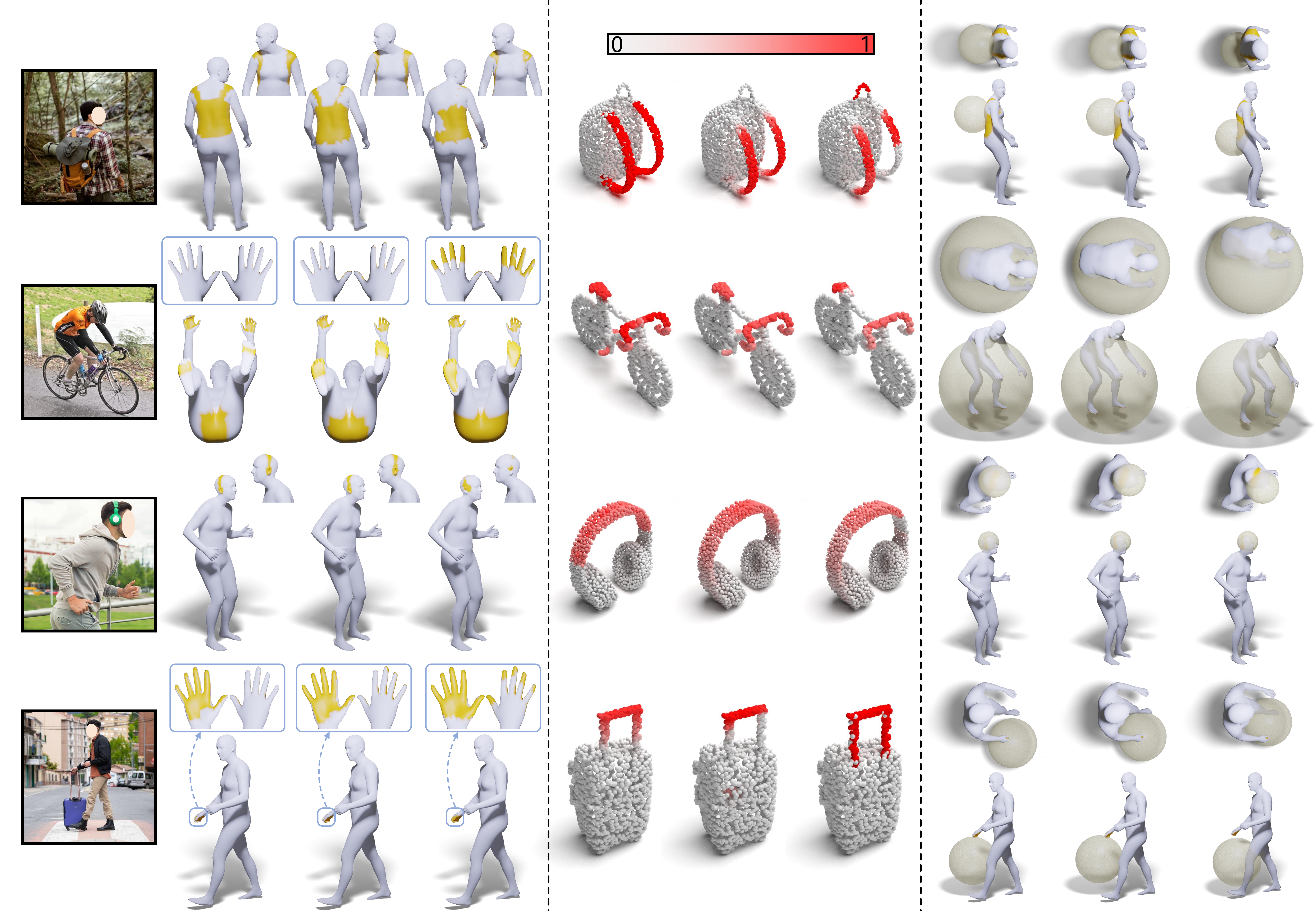}
        \put(16,68.5){{\textbf{\footnotesize GT}}}
        \put(24.5,68.5){{\textbf{\footnotesize Ours}}}
        \put(32.5,68.5){{\textbf{\footnotesize DECO} \cite{tripathi2023deco}}}

        \put(45,68.5){{\textbf{\footnotesize GT}}}
        \put(54.5,68.5){{\textbf{\footnotesize Ours}}}
        \put(62.5,68.5){{\textbf{\footnotesize IAG} \cite{Yang_2023_ICCV}}}

        \put(74,68.5){{\textbf{\footnotesize GT}}}
        \put(83,68.5){{\textbf{\footnotesize Ours}}}
        \put(90,68.5){{\textbf{\footnotesize PopUp} \cite{petrov2023popup}}}
        \put(97.5,63){\rotatebox{90}{\textbf{\footnotesize View 1}}}
        \put(97.5,56){\rotatebox{90}{\textbf{\footnotesize View 2}}}

        \put(26,-1){{\textbf{\footnotesize (a)}}}
        \put(55,-1){{\textbf{\footnotesize (b)}}}
        \put(85,-1){{\textbf{\footnotesize (c)}}}

        \put(54.5,51){{\textbf{\footnotesize Carry}}}
        \put(55,34.5){{\textbf{\footnotesize Ride}}}
        \put(55,19){{\textbf{\footnotesize Wear}}}
        \put(55,1){{\textbf{\footnotesize Pull}}}

        \put(3.5,51.5){{\textbf{\footnotesize Backpack}}}
        \put(4.4,35.3){{\textbf{\footnotesize Bicycle}}}
        \put(3.5,19){{\textbf{\footnotesize Earphone}}}
        \put(4,2.8){{\textbf{\footnotesize Suitcase}}}
        
	\end{overpic}
	\caption{\textbf{Visualization Results.} \textbf{(a)} Results of the estimated human vertices in contact with objects, the estimated contact vertices are shown in \textcolor[rgb]{0.88,0.82,0.36}{yellow}. \textbf{(b)} The anticipations of 3D object affordance, the depth of \textcolor[rgb]{0.83,0.24,0.21}{red} represents the probability of anticipated affordance. \textbf{(c)} Two views of the predicted spatial relation, translucent spheres are object proxies. Please zoom in for a better visualization.}
 \label{Fig:mainresults}
\end{figure*}

\textbf{Statistical Analysis.} The quantity of images and 3D instances for each object category in the 3DIR is shown in Fig. \ref{Fig:dataset} (a). For contact annotation, we count its proportion within 24 human parts defined on the SMPL \cite{loper2023smpl}. Moreover, we visualize the distribution of contact annotation for several HOIs, where the deeper color indicates more contact annotations at the vertex, as shown in Fig. \ref{Fig:dataset} (b). Fig. \ref{Fig:dataset} (c) demonstrates the proportion of annotated affordance regions to the entire object geometry, encompassing the mean of each category and the distribution of all instances for several categories. As can be seen, there are differences in distinct objects with the same affordance, as well as variations in distinct affordances of the same object. For spatial relation annotation, we count the mean distance between annotated centers and human pelvis joints, and project the spatial direction of annotated centers onto a fixed-radius sphere with the pelvis joint as the center. Fig. \ref{Fig:dataset} (d) shows several cases and the distribution of mean distances.
\section{Experiment}
\subsection{Benchmark Setting}
We refer to methods that anticipate each interaction element \cite{tripathi2023deco, Huang:CVPR:2022, Yang_2023_ICCV, nagarajan2019grounded} to thoroughly benchmark the 3DIR. For the training, in addition to the training data in 3DIR, we select another 5k data with low redundancy in BEHAVE \cite{bhatnagar22behave}. Evaluation is conducted on the test set of 3DIR. The baseline model adopts a multitask-like framework, directly utilizing three branches to make anticipations. Besides, we compare LEMON with advanced methods that anticipate respective elements. Plus, we also conduct evaluation experiments on DAMON \cite{tripathi2023deco}, BEHAVE \cite{bhatnagar22behave}, and PIAD \cite{Yang_2023_ICCV}, the results and implementation details are in the Sup. Mat.  

\subsection{Comparison Results}
The comparison results of evaluation metrics are presented in Tab. \ref{table:main_results}. The baseline model, which does not capture correlations between interaction counterparts, indicates that directly anticipating these elements through multiple branches yields poor results. Our method outperforms comparative methods across all metrics for respective elements, demonstrating that leveraging correlations between humans and objects indeed benefits the comprehension of interaction relation, and the anticipation of interaction elements. It seems to be ``the best of both worlds''. Furthermore, we conduct a visual comparison of methods with higher evaluation metrics, as shown in Fig. \ref{Fig:mainresults}. The results showcase multiple objects interacting with distinct human parts. As can be seen, our anticipations are more precise, for some uncertain regions that are not visible in images (\eg, the first and third rows), LEMON could also anticipate plausible results. This is attributed to the modeling of correlations between geometries, which compensates for missing features in invisible regions of the image.

\subsection{Ablation Study}
We conduct a thorough ablation study to validate the effectiveness of the model design. Tab. \ref{table:ablation} (a) reports the model performance without modeling the semantic intention and the geometric correlation, demonstrating their impacts on the model performance. Moreover, the absence of constraining semantic consistency hinders the modeling of intentions, while not introducing curvatures affects the extraction of geometric correlations, resulting in a decrease in model performance. Both results are detailed in Tab. \ref{table:ablation} (a). Besides, we test the impact of $\phi_{c}$ and $\mathbf{\mathcal{L}}_{pr}$ (Eq. \ref{Eq:6}) on spatial relation prediction, shown in Tab. \ref{table:ablation} (b). Notably, removing the extraction of $\phi_{a}$ in $f_{\theta}$ (Eq. \ref{Eq:3}) also affects the performance. This is attributed to the interrelation between $\phi_{c}$ and $\phi_{a}$, the absence of $\phi_{a}$ impacts the representation of $\phi_{c}$, and subsequently influencing the final spatial prediction.

\begin{figure}
    \centering
    \small
    \begin{overpic}[width=0.94\linewidth]{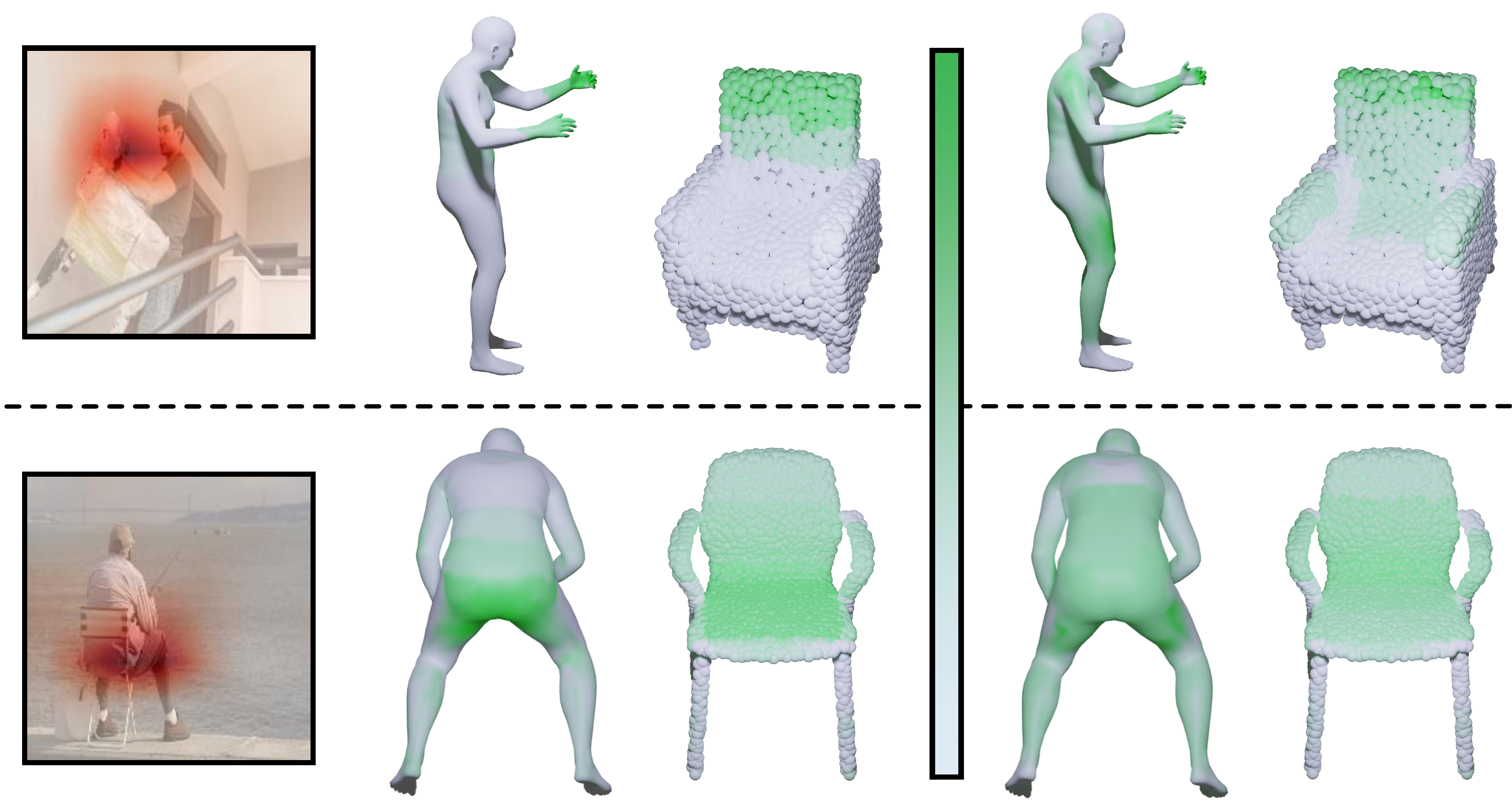}
    \put(32,53.5){\textbf{$\bm{w}$ intention}}
    \put(72,53.5){\textbf{$\bm{w/o}$ intention}}
    \put(33,-2){\textbf{$\bm{w}$ geometry}}
    \put(71,-2){\textbf{$\bm{w/o}$ geometry}}

    \put(7.4,28.1){\textbf{\footnotesize Move}}
    \put(9.3,-0.2){\textbf{\footnotesize Sit}}
    
    \put(59.5,51.5){\textbf{\footnotesize high}}
    \put(60,-0.5){\textbf{\footnotesize low}}
    \end{overpic}
    \caption{\textbf{Visual Attention.} Attention heatmaps on geometries, in cases with ($\bm{w}$) and without ($\bm{w/o}$) the extraction of intention and geometry affinity. Images are also shown with attention heatmaps.}
    \label{fig:ablation}
    \vspace{-0.21cm}
\end{figure}

To further illustrate the effectiveness of capturing the intention and geometry affinity, we visualize attention maps on the human and object geometries when removing one of them, as shown in Fig. \ref{fig:ablation}. The first row indicates without capturing intention affinity, solely relying on geometries leads the model to focus on multiple candidate regions, differing from attention regions in the image. Furthermore, the second row demonstrates that while semantic intention guides the model to identify approximate candidate regions, modeling geometry affinity further locates corresponding regions most related to the interaction of the geometries.

\begin{table}[t]
\footnotesize
  \renewcommand{\arraystretch}{1.}
  \renewcommand{\tabcolsep}{4.pt}
  \caption{{\textbf{Ablation studies.} \textbf{(a)} Performance when not modeling intention representations (intent.), geometric correlations (geom.), and not introducing $\varphi$ (Eq. \ref{Eq:1}), curvatures (cur.). \textbf{(b)} The impact of several factors on spatial relation prediction. \ding{55} means without.}}
\label{table:ablation}
\begin{subtable}[t]{0.8\linewidth}
\begin{tabular}{c|c|cccc}
\toprule
\textbf{Metrics}   & \cellcolor{mygray}\textbf{Ours} & \textbf{\ding{55} intent.} & \textbf{\ding{55} $\varphi$} & \textbf{\ding{55} geom.} & \textbf{\ding{55} cur.} \\ \midrule
\textbf{Precision}     & \cellcolor{mygray}0.78             & 0.71       &   0.74      & 0.68                        & 0.75\\
\textbf{Recall}    & \cellcolor{mygray}0.82             & 0.73        &    0.79    & 0.70                        & 0.78\\
\textbf{F1}        & \cellcolor{mygray}0.78             & 0.70        &    0.74    & 0.67                        & 0.75\\
\textbf{geo. (cm)}      & \cellcolor{mygray}7.55            & 11.87       &   10.26     & 14.87                       & 9.13\\ \cmidrule{1-6}
\textbf{AUC}       & \cellcolor{mygray}88.51            & 85.87          &  86.66   & 83.23                       & 86.98\\
\textbf{aIOU}      & \cellcolor{mygray}41.34            & 38.19        &   40.03    & 37.13                       & 39.58\\
\textbf{SIM}       & \cellcolor{mygray}0.64            & 0.58         &   0.60    & 0.55                        & 0.62\\ \cmidrule{1-6}
\textbf{MSE}       & \cellcolor{mygray}0.010            & 0.027       &   0.022      & 0.031                       & 0.018\\ \bottomrule
\end{tabular}
\caption{}
\end{subtable}
\begin{subtable}[t]{0.18\linewidth}
\centering
\begin{tabular}{c|c}
\toprule
   & \textbf{MSE} \\ \midrule
\rotatebox[origin=c]{90}{\textbf{\ding{55} $\phi_{c}$}} & 0.037 \\ \midrule
\rotatebox[origin=c]{90}{\textbf{\ding{55} $\mathbf{\mathcal{L}}_{pr}$}} & 0.024 \\ \midrule
\rotatebox[origin=c]{90}{\textbf{\ding{55} $\phi_{a}$}} & 0.019\\
\bottomrule
\end{tabular}
\caption{}
\end{subtable}
\vspace{-20pt}
\end{table}

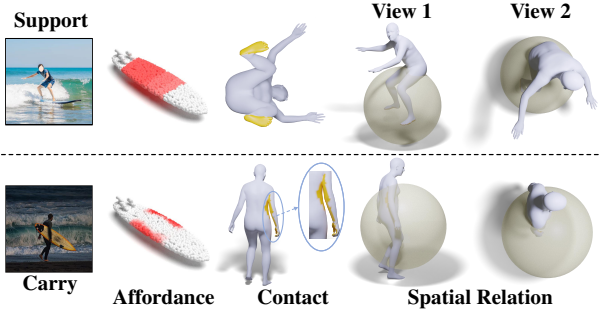
\begin{figure}
    \centering
    \small
    \begin{overpic}[width=0.96\linewidth]{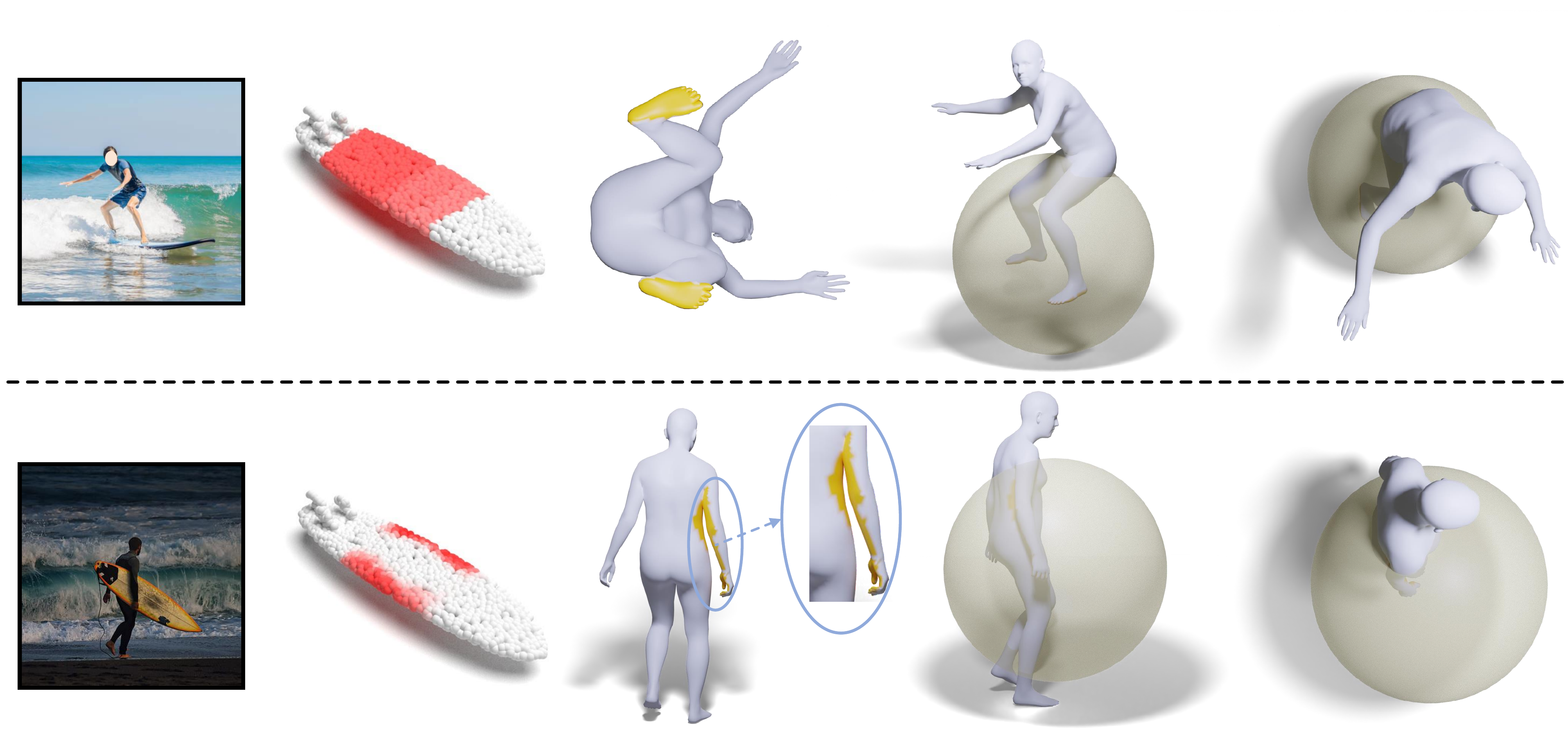}
    \put(2.6,43.5){\textbf{\footnotesize Support}}
    \put(62,45){\textbf{\footnotesize View 1}}
    \put(85,45){\textbf{\footnotesize View 2}}
    
    \put(4,-0.2){\textbf{\footnotesize Carry}}
    \put(19,-2.5){\textbf{\footnotesize Affordance}}
    \put(43,-2.5){\textbf{\footnotesize Contact}}
    \put(68,-2.5){\textbf{\footnotesize Spatial Relation}}
    \end{overpic}
    \caption{\textbf{Multiple Interactions.} Distinct anticipations when interactions are different with the same object.}
    \vspace{2.5pt}
    \label{fig:multiaction}
\end{figure}

\subsection{Performance Analysis}
\textbf{Multiple Interactions.} To validate whether the model reasons interaction elements based on the understanding of interaction relation. We employ the model to infer different interactions between humans and the same object, as demonstrated in Fig. \ref{fig:multiaction}. The results showcase that, for the same object, when interactions are different, the model outputs distinct results, encompassing human contact, object affordance, and spatial relation, all of which align with the interaction relation depicted in the image. This indicates that the model infers interaction elements according to the interaction relation rather than relying on direct mapping solely driven by the categories of human-object pairs.

\begin{figure}
    \centering
    \small
    \begin{overpic}[width=0.95\linewidth]{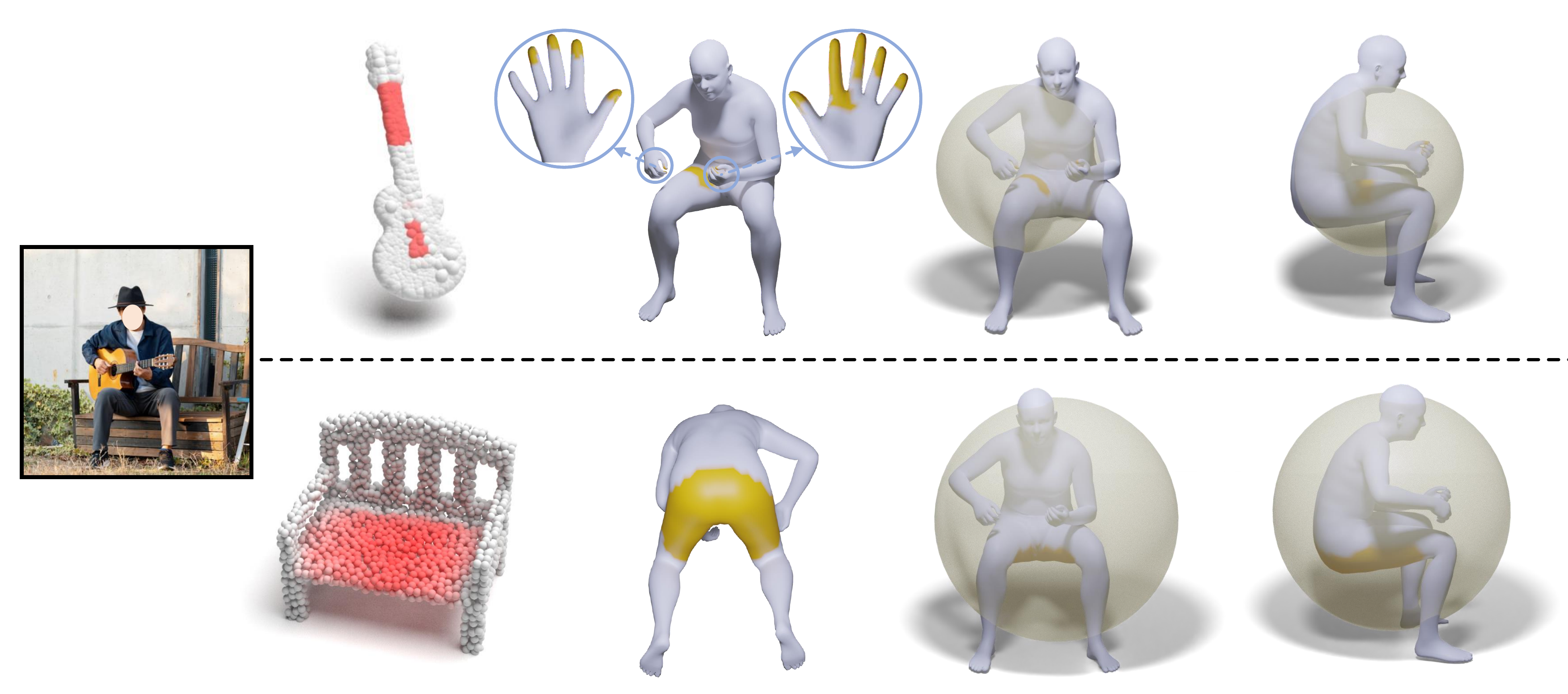}
    \put(3.5,35){\textbf{\footnotesize Guitar}}
    \put(5.5,32){\textbf{\footnotesize Play}}
    \put(4.8,8){\textbf{\footnotesize Chair}}
    \put(6.8,5){\textbf{\footnotesize Sit}}
    
    \put(17,-2.5){\textbf{\footnotesize Affordance}}
    \put(41,-2.5){\textbf{\footnotesize Contact}}
    \put(71,-2.5){\textbf{\footnotesize Spatial Rel.}}
    \end{overpic}
    \caption{\textbf{Multiple Objects.} The anticipations for multiple objects with different human-object interactions.}
    \label{fig:multicontact}
    \vspace{-0.1cm}
\end{figure}

\noindent\textbf{Multiple Objects.} In scenarios where distinct objects are involved in concurrent interactions with humans, the model should indeed possess the ability to comprehend the interaction relation with different objects. We give an experiment pertaining to this property. As shown in Fig. \ref{fig:multicontact}, when reasoning with different objects, the human contact, object affordance, and spatial relation are distinct, while all anticipations are plausible. This is attributed to the collaboration of the semantic interaction intention and geometric correlation, ensuring that the model could explicitly capture interaction intentions with different objects and reveal regions related to the interaction on corresponding geometries.

\begin{figure}
    \centering
    \small
    \begin{overpic}[width=0.95\linewidth]{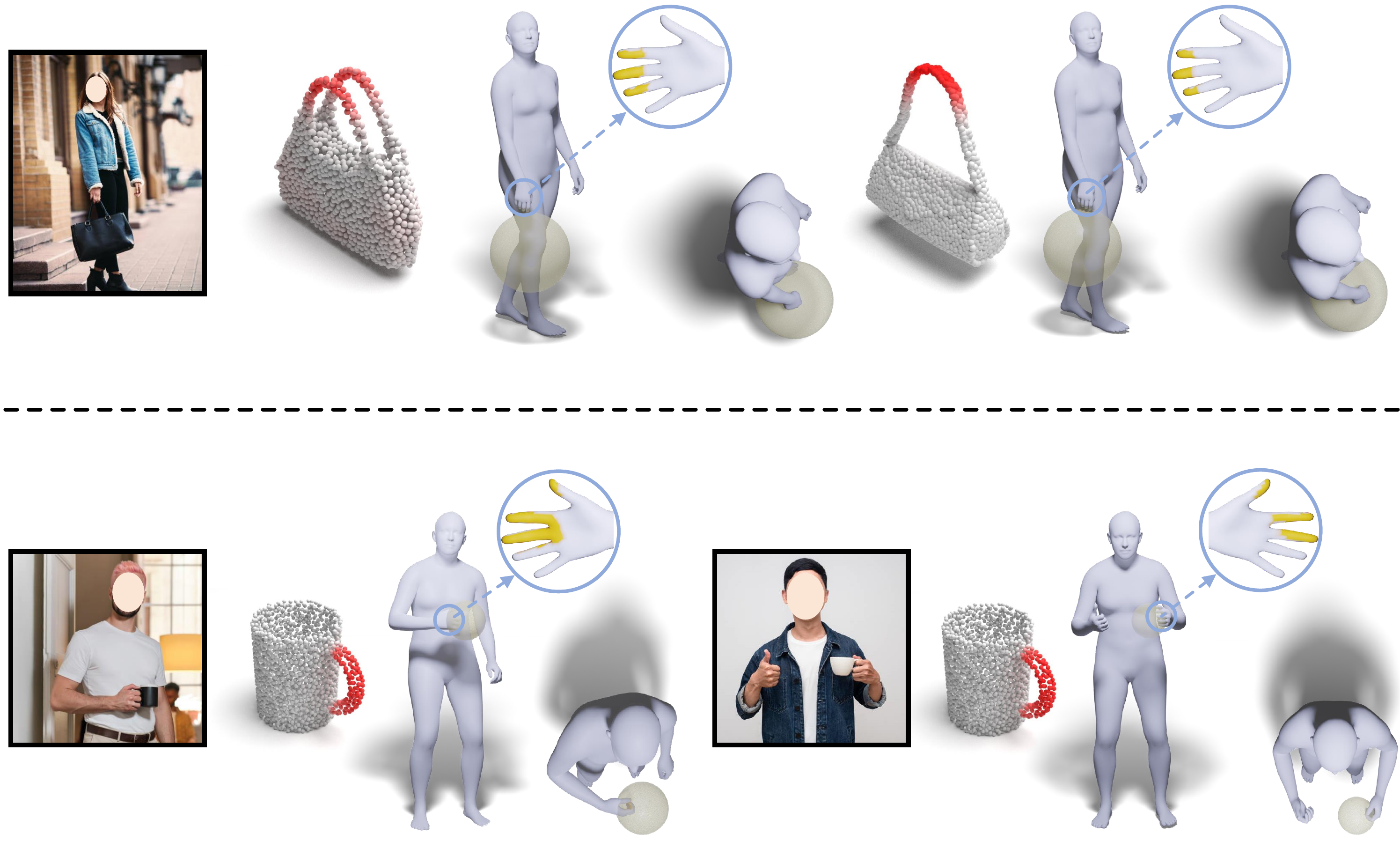}
    \put(16,33.5){\textbf{\footnotesize Affordance}}
    \put(59,33.5){\textbf{\footnotesize Affordance}}
    \put(42,62.5){\textbf{\footnotesize Contact}}
    \put(82,62.5){\textbf{\footnotesize Contact}}
    \put(38.5,33.5){\textbf{\footnotesize Spatial Rel.}}
    \put(79,33.5){\textbf{\footnotesize Spatial Rel.}}
    \put(4,59.5){\textbf{\footnotesize Lift}}
    
    \put(6,33.5){\textbf{\footnotesize (a)}}
    \put(6,-2){\textbf{\footnotesize (b)}}
    
    \put(3,24){\textbf{\footnotesize Grasp}}
    \put(53,24){\textbf{\footnotesize Grasp}}
    \put(16,-2){\textbf{\footnotesize Afford.}}
    \put(66,-2){\textbf{\footnotesize Afford.}}
    \put(34,29){\textbf{\footnotesize Contact}}
    \put(83.5,29){\textbf{\footnotesize Contact}}
    \put(30.5,-2){\textbf{\footnotesize Spatial Rel.}}
    \put(81,-2){\textbf{\footnotesize Spatial Rel.}}
    \end{overpic}
    \caption{\textbf{Multiple Instances.} \textbf{(a)} Different object instances $w.r.t.$ the same image and human geometry. \textbf{(b)} Different images and human instances $w.r.t.$ the same object.}
    \vspace{4pt}
    \label{fig:one2multi}
\end{figure}
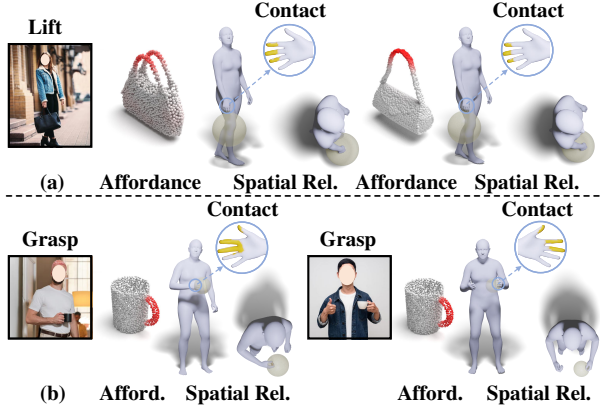

\noindent\textbf{Multiple Instances.} We conduct an experiment to validate the model's understanding of interaction relation across various instances, assessing its generalization and robustness. Specifically, distinct object instances are utilized to infer with the same image and human geometry, shown in Fig. \ref{fig:one2multi} (a). As can be seen, even though the contact and spatial relation are slightly different, the results are consistent with the interaction shown in the image, and object affordances are correctly anticipated. Additionally, the same object instance is utilized to infer with different human geometries and images that contain similar human-object interactions, as demonstrated in Fig. \ref{fig:one2multi} (b). These results indicate that the model could generalize the understanding of interaction relation to different human and object instances.

\section{Conclusion}

We propose leveraging both counterparts of the HOI to jointly anticipate human contact, object affordance, and human-object spatial relation in 3D space. It holds the potential to drive embodied intelligence to learn from interactions and assist the interaction modeling in graphics. With the provided dataset 3DIR containing paired HOI data and multiple annotations, we train LEMON, a novel model that harnesses the interaction intention and geometric correlation between humans and objects to capture the affinity, eliminating interaction uncertainties and anticipating plausible interaction elements. Extensive experiments demonstrate the effectiveness and superiority of LEMON. We believe this work offers fresh insights and paves a new way for 3D human-object interaction understanding.

\noindent\textbf{Limitations and Future Work.} In the current formulation, LEMON needs the pre-inferred human mesh as input. In the future, we plan to integrate HMR into the entire framework, freeing input constraints required for inference, and taking 3D interaction elements to improve the accuracy of HMR. Plus, another interesting direction is to leverage multi-modal methods \cite{girdhar2023imagebind, zhang2022pointclip}, boosting the interaction relation understanding from other sources \eg, text, audio.

\noindent\textbf{Acknowledgments}. This work is supported by the National Natural Science Foundation of China (NSFC) under Grants 62225207 and 62306295.

\section{Appendix}
\section*{A. Implementation Details}
\addcontentsline{toc}{section}{\textcolor[rgb]{0,0,0}{A. Implementation Details}}

\quad In this section, we clarify details about the methods, experiments, and training settings.

\subsection*{A.1. Method Details}
\addcontentsline{toc}{subsection}{\textcolor[rgb]{0,0,0}{A.1. Method Details}}

\quad We provide a table to display dimensions and meanings of tensors in the LEMON pipeline, shown in Tab. \ref{tab:tensor}. The input images are resized to $224\times$224 and a pre-trained HRNet is taken as the extractor. We add human and object masks in the training to enable the model to focus efficiently on the interaction between the human and object. The image extractor outputs the image feature with the shape $\mathbb{R}^{2048 \times 7 \times 7}$. A $1\times1$ convolutional layer is used to reduce the feature dimension and the feature is flattened to $\mathbf{F}_{i} \in \mathbb{R}^{768 \times 49}$. The number of object point clouds is $2048$, the same setting as 3D-AffordanceNet \cite{deng20213d} and IAG-Net \cite{Yang_2023_ICCV}. For humans, the vertices of SMPL-H \cite{MANO:SIGGRAPHASIA:2017} are regarded as the input, raw vertices possess the shape $\mathbb{R}^{6890 \times 3}$. We sample it to $\mathbb{R}^{1723 \times 3}$ through the script in COMA \cite{COMA:ECCV18}, which is also utilized in BSTRO \cite{Huang:CVPR:2022}. Note that we uniformly use $N_{h}$ to represent the number of vertices in the main paper for simplification. Actually, $N_{h}$ in $H$ and $\bar{\phi}_c$ are $6890$, while in other features are $1723$. DGCNN is taken as the backbone network for extracting point-wise features of the human and object and output the $\mathbf{F}_{o} \in \mathbb{R}^{768 \times 2048}, \mathbf{F}_{h} \in \mathbb{R}^{768 \times 1723}$.

The tokens $\mathbf{T}_{o}, \mathbf{T}_{h} \in \mathbb{R}^{768 \times 1}$ are utilized to represent interaction intentions of geometries. They are concatenated with $\mathbf{F}_{o}, \mathbf{F}_{h}$, and get the feature sequence $\mathbf{F}_{to} \in \mathbb{R}^{768 \times 2049}, \mathbf{F}_{th} \in \mathbb{R}^{768 \times 1724}$. Then, the multi-branch attention $f_{\delta}$ is performed on $\mathbf{F}_{i}$ and $\mathbf{F}_{to}, \mathbf{F}_{th}$. $\mathbf{F}_{i}$ serves as the shared key and value, $\mathbf{F}_{to}, \mathbf{F}_{th}$ serve as queries in two branches. $f_{\delta}$ has $12$ heads and each head with the dimension of $64$. After $f_{\delta}$, $\mathbf{F}_{to}, \mathbf{F}_{th}$ are updated to $\bar{\mathbf{F}}_{to}, \bar{\mathbf{F}}_{th}$ with the same shape, which are then split to $\bar{\mathbf{F}}_{o} \in \mathbb{R}^{768 \times 2048}, \bar{\mathbf{T}}_{o} \in \mathbb{R}^{768 \times 1}$ and $\bar{\mathbf{F}}_{h} \in \mathbb{R}^{768 \times 1723}, \bar{\mathbf{T}}_{h} \in \mathbb{R}^{768 \times 1}$.

\begin{table}[]
  \centering
  \renewcommand{\arraystretch}{1.}
  \renewcommand{\tabcolsep}{8pt}
    \footnotesize
   \caption{\textbf{Tensors.} The dimension and meaning of the tensors in the LEMON pipeline.}
   \label{tab:tensor}
\begin{tabular}{c|c|c}
\toprule
\textbf{Tensor}    & \textbf{Dimension} & \textbf{Meaning} \\ \midrule
$\mathbf{F}_{i}$       & $768 \times 49$           & image feature            \\
$\mathbf{F}_{o}, \bar{\mathbf{F}}_{o}$    & $768 \times 2048$           & geometric feature of the object    \\
$\mathbf{F}_{h}, \bar{\mathbf{F}}_{h}$        & $768 \times 1723$                 & geometric feature of the human           \\
$\mathbf{T}_{o}, \bar{\mathbf{T}}_{o}$  & $768 \times 1$   & intention tokens of object geometry            \\
$\mathbf{T}_{h}, \bar{\mathbf{T}}_{h}$    & $768 \times 1$           & intention tokens of human geometry    \\
$\mathbf{F}_{to}$  & $768 \times 2049$   & concatenation of $\mathbf{F}_{o}, \mathbf{T}_{o}$     \\
$\bar{\mathbf{F}}_{to}$     &  $768 \times 2049$      & concatenation of $\bar{\mathbf{F}}_{o}, \bar{\mathbf{T}}_{o}$           \\
$\mathbf{F}_{th}$       & $768 \times 1724$          & concatenation of $\mathbf{F}_{h}, \mathbf{T}_{h}$            \\
$\bar{\mathbf{F}}_{th}$   & $768 \times 1724$               & concatenation of $\bar{\mathbf{F}}_{h}, \bar{\mathbf{T}}_{h}$           \\
 $C_o$           &  $1 \times 2048$                & curvature of the object geometry          \\
$C_h$        & $1 \times 1723$             & curvature of the human geometry          \\
$\bar{C}_o$        & $768 \times 2048$             & curvature feature of the object         \\
 $\bar{C}_h$ &     $768 \times 1723$  &  curvature feature of the human \\
  $\bar{\mathbf{F}}_{co}$ &     $768 \times 2048$  &  geometric feature with curvature \\
$\bar{\mathbf{F}}_{ch}$        & $768 \times 1723$   & geometric feature with curvature          \\
$\mathbf{T}_{sp}$        & $768 \times 3$   & spatial token sequence         \\
$\phi_{a}$      & $768 \times 2048$                & object affordance representation          \\ 
$\phi_{c}$      & $768 \times 1723$             & human contact representation\\
$\phi_{p}$      & $768 \times 3$                & object spatial representation\\
\bottomrule
\end{tabular}
\end{table}

\begin{figure}
    \centering
    \begin{overpic}[width=0.9\linewidth]{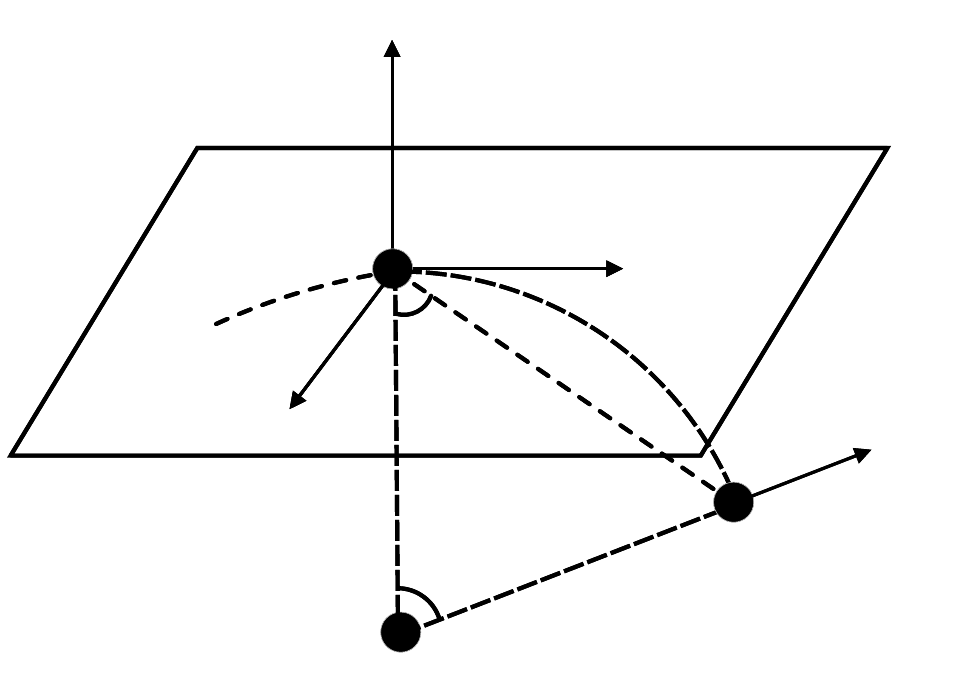}
    \put(42,47){p}
    \put(43,35){\textbf{$\mathbf{\theta}$}}
    \put(43.5,13){$\mathbf{\beta}$}
    \put(39,0){$\mathbf{O}$}
    \put(73,14){$m_i$}
    \put(91,25){$M_i$}
    \put(27.5,25){$X$}
    \put(65,42){$Y$}
    \put(38.5,68){$N$}
    
    \end{overpic}
    \caption{The local coordinates system L with the triangle defined by the osculating circle neighbor point of $p$.}
    \label{fig:sup_curvature}
\end{figure}
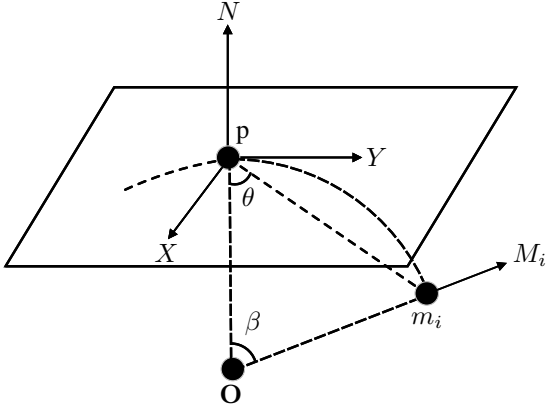

$\bar{\mathbf{F}}_{o}, \bar{\mathbf{F}}_{h}$ and $\bar{\mathbf{T}}_{o}, \bar{\mathbf{T}}_{h}$ are utilized to model the geometric correlation. Firstly, we introduce the method to calculate the geometric curvature. Shown in Fig. \ref{fig:sup_curvature}, for each point $p$ in the point cloud, the neighbor points are utilized to estimate its normal curvature. Assume $p$ has $n$ neighbor points, and let $m_{i}$ be the $i\text{-}th$ neighbor point. The normal vector corresponding to $m_{i}$ is $\vec{M}_i$. Define ${p, \vec{X}_i, \vec{Y}_i, \vec{N}_i}$ be an orthogonal coordinates system, which is the local coordinates $L$ at point $p$. $\vec{N}_i$ is the normal vector of $p$, $\vec{X}_i, \vec{Y}_i$ are orthogonal unit vectors. In $L$, coordinates could be formulated as: $p (0,0,0), m_{i} (x_i, y_i, z_i), \vec{M}_i (n_{x,i}, n_{y,i}, n_{z,i})$. Then, the normal curvature $C^{i}_{n}$ of $p$ could be calculated with an osculating circle passing through point $p$ and $m_{i}$, which could be expressed as:
\begin{equation}
\small
\begin{aligned}
    C_n^i=&-\frac{\sin \beta}{\left|p m_i\right| \sin \theta} \approx-\frac{n_{x y}}{\sqrt{n_{x y}^2+n_z^2} \sqrt{x_i^2+y_i^2}}, \\
    &where \quad n_{x y}=\frac{x_i n_{x, i}+y_i n_{y, i}}{\sqrt{x_i^2+y_i^2}}, n_z=n_{z, i},
\end{aligned}
\end{equation}
where $\theta$ is the included angle between vectors $\text{-}\vec{N}_i$ and $\vec{pm_{i}}$, $\beta$ is between vectors $\vec{N}$ and $\vec{M}_i$. Taking this method to obtain the curvature $C_o, C_h$. For the convenience of future research, we store these curvatures for direct use by other researchers. As described in the main paper, $C_o, C_h$ are encoded into high dimension, and the cross-attention $f_m$ is mutually performed on them. $f_m$ also possesses $12$ heads and each head with the dimension of $64$. The fusion layer $f$ (Sec. 3.2 in the main paper) is the $1 \times 1$ convolution layer, which reduces the fused feature dimension to $768$ for subsequent calculations.

To model the object's spatial representation, the token sequence $\mathbf{T}_{sp} \in \mathbb{R}^{768 \times 3}$ is concatenated with the semantic token $\bar{\mathbf{T}}_{o}$ and the global geometric feature of the object. With the addition of positional encoding, the concatenate feature is utilized to query the corresponding feature of the human through a cross-attention layer $f_{\rho}$. The global geometric features are obtained by max-pooling $\bar{\mathbf{F}}_{co}, \bar{\mathbf{F}}_{ch}$, $f_{\rho}$ has the same architecture with $f_m$.

The decoder has three heads to project the output of human contact $\bar{\phi}_c$, object affordance $\bar{\phi}_a$, and object center position $\bar{\phi}_p$. Each head is composed of a linear layer, a batch-normalization layer, and an activation layer. $\phi_a$ is projected to $\bar{\phi}_a \in \mathbb{R}^{2048 \times 1}$, and $\phi_p$ is projected to $\bar{\phi}_p \in \mathbb{R}^{3}$. For the contact feature, it is first projected to $\mathbb{R}^{1723 \times 1}$ and then up-project to $\bar{\phi}_c \in \mathbb{R}^{6890 \times 1}$ by another linear layer. We give the formulation of $\mathbf{\mathcal{L}}_{p}, \mathbf{\mathcal{L}}_{s}$ in the main paper. Here, we also provide the formulation of $\mathbf{\mathcal{L}}_{c}, \mathbf{\mathcal{L}}_{a}$, which are the same, expressed as:
\begin{equation}
\small
\begin{aligned}
    \mathcal{L}_{c}, \mathcal{L}_{a} = &1-\frac{\sum_j^N y x+\epsilon}{\sum_j^N y+x+\epsilon}
    -\frac{\sum_j^N\left(1-y\right)\left(1-x\right)+\epsilon}{\sum_j^N 2-y-x+\epsilon}\\
    & +\frac{1}{N}\sum_j^N[-\left(1-\alpha)(1-y\right) x^\gamma \log \left(1-x\right)\\
    &-\alpha y\left(1-x\right)^\gamma \log \left(x\right)],
\end{aligned}
\end{equation}
where $N$ indicates the number of points within each geometry, $x$ is the prediction, $y$ is the ground truth, $\epsilon$ is set to $1e$-$6$, $\alpha, \gamma$ are set to $0.25$ and $2$ respectively.

\subsection*{A.2. Benchmark Details}
\addcontentsline{toc}{subsection}{\textcolor[rgb]{0,0,0}{A.2. Benchmark Details}}

\textbf{Evaluation Metrics.} We refer to methods that estimate each interaction element to benchmark the 3DIR \cite{tripathi2023deco, Huang:CVPR:2022, deng20213d, Yang_2023_ICCV, petrov2023popup}. Specifically, the Precision, Recall, F1 score, and geodesic distance are utilized to evaluate the contact estimation. AUC, aIOU, and SIM are utilized to evaluate the anticipation of object affordance. MSE is taken to evaluate the prediction of objects' spatial positions. The details of these metrics are as follows:

\begin{itemize}

\item [$\bm{-}$] \textbf{Precision, Recall, F1 }\cite{goutte2005probabilistic}: Precision is the ratio of correctly predicted positive observations to the total predicted positives, measures the accuracy of the positive predictions made by a model. Recall is the ratio of correctly predicted positive observations to all observations in the actual class and measures the ability of a model to capture all the positive instances. F1-score is the harmonic mean of Precision and Recall. It provides a balance between Precision and Recall, making it a suitable metric when there is an imbalance between classes. They could be formulated as:
\begin{equation}
\small
\begin{aligned}
    &Precision = \frac{TP}{TP+FP},\\
    &Recall = \frac{TP}{TP+FN},\\
    &F1 = \frac{2 \cdot Precision \cdot Recall}{Precision + Recall},
\end{aligned}
\end{equation}
where $TP$, $FP$, and $FN$ denote the true positive, false positive, and false negative counts, respectively.

\item [$\bm{-}$] \textbf{geodesic distance }\cite{Huang:CVPR:2022}: The geodesic distance is utilized to translate the count-based scores to errors in metric space. For each vertex predicted in contact, its shortest geodesic distance to a ground-truth vertex in contact is calculated. If it is a true positive, this distance is zero. If not, this distance indicates the amount of prediction error along the body.

\item [$\bm{-}$] \textbf{AUC }\cite{lobo2008auc}: The Area under the ROC curve, referred to as AUC, is the most widely used metric for evaluating saliency maps. The saliency map is treated as a binary classifier of fixations at various threshold values (level sets), and a ROC curve is swept out by measuring the true and false positive rates under each binary classifier.

\item [$\bm{-}$] \textbf{aIOU }\cite{rahman2016optimizing}: IoU is the most commonly used metric for comparing the similarity between two arbitrary shapes. The IoU measure gives the similarity between the predicted region and the ground-truth region, and is defined as the size of the intersection divided by the union of the two regions. It can be formulated as:

\begin{equation}
\small
    IoU = \frac{TP}{TP+FP+FN},
\end{equation}
where $TP$, $FP$, and $FN$ denote the true positive, false positive, and false negative counts, respectively.

\item [$\bm{-}$] \textbf{SIM }\cite{swain1991color}: The similarity metric (SIM) measures the similarity between the prediction map and the ground truth map. Given a prediction map $P$ and a continuous ground truth map $Q^{D}$, $SIM(\cdot)$ is computed as the sum of the minimum values at each element, after normalizing the input maps:
\begin{equation}
\small
\begin{split}
   &SIM (P,Q^{D})=\sum_{i}min(P_{i},Q_{i}^{D}),\\
    & where\quad \sum_{i}P_{i}=\sum_{i}Q_{i}^{D}=1.
\end{split}
\end{equation}

\item [$\bm{-}$] \textbf{MSE }\cite{wang2009mean}: The Mean Squared Error (MSE) is a measure of the average squared difference between the predicted and actual values in a regression problem. MSE quantifies the average squared difference between the predicted and actual values. It penalizes larger errors more heavily than smaller errors, making it sensitive to outliers. Lower MSE values indicate better model performance in terms of regression accuracy. It is formulated as dividing the total error by $n$:

\begin{equation}
\small
\mathrm{MSE}=\frac{1}{n} \sum_{i=1}^n\left|\hat{y}-y\right|_{2},
\end{equation}
where $y$ is the prediction, $\hat{y}$ is the ground truth.
\end{itemize}

\noindent\textbf{Comparison Methods.} Here, we provide an elucidation of the implementation for the methods employed for comparison in the experiment.
\begin{itemize}
\item [$\bm{-}$] \textbf{Baseline}: The baseline model simply takes modality-specific backbones to extract respective features of the image and geometries. Then, it decodes these features separately to obtain the outputs through three branches. This verifies the performance when directly treating this task as a multi-task regression.

\item [$\bm{-}$] \textbf{BSTRO }\cite{Huang:CVPR:2022}: BSTRO takes the HRNet as the image backbone and concat the human vertices with the extracted image feature, then it utilizes a multi-layer transformer to estimate the contact vertex. We retain the network architecture while introducing a modification: the vertex of human mesh in BSTRO is downsampled to $431$. However, we find that $1723$ vertices achieve superior results through experiments. Consequently, during the training, the vertices are downsampled to $1723$, the same with LEMON.

\item [$\bm{-}$] \textbf{DECO }\cite{tripathi2023deco}: DECO possesses the scene and part context branch to parse the semantics in images, thus facilitating the estimation of human contact. We follow the authors' instructions, taking the Mask2Former \cite{cheng2021mask2former} to create scene segmentation maps for images in 3DIR. Using the scene and contact branches to train the DECO on 3DIR.

\item [$\bm{-}$] \textbf{3D-AffordanceNet }\cite{deng20213d}: 3D-AffordanceNet directly utilizes the DGCNN or PointNet++ to extract per-point features and decodes them to the affordance representation. It tends to anticipate all affordances of the object and may not be consistent with the object affordance in the image. To this end, we slightly modify its structure, taking a cross-attention to update the geometric feature, with the point feature as the query and the image feature as key and value. The affordance representation is obtained by decoding the updated geometric features.

\item [$\bm{-}$] \textbf{IAG-Net }\cite{Yang_2023_ICCV}: IAG-Net anticipates the object affordance of objects by establishing the correlations between interaction contents in the image and the geometric features of the object point cloud. We directly utilize the original architecture of IAG-Net to train on the 3DIR. It is worth noting that the training of IAG-Net needs the bounding boxes of the interactive subject and object. We obtain the bounding box by taking the positions of pixels with the smallest and largest in horizontal and vertical coordinates from mask annotations in 3DIR.

\item [$\bm{-}$] \textbf{DJ-RN }\cite{li2020detailed}: DJ-RN defines the radius for various types of objects, using a sphere to represent the object. Which lifts the spatial relation in images to the 3D space by leveraging the bounding boxes of humans and objects in pixel space and the defined radius. We use its official code to infer the 3DIR data. Additionally, we make minor adjustments to the radius of objects to match the 3D objects in 3DIR (minimal impact on the results).

\item [$\bm{-}$] \textbf{Object pop-up }\cite{petrov2023popup}: This method takes vertices of a posed human as the input, anticipating what objects could interact with the human and the object's spatial position. Since we benchmark it as a comparative method for predicting spatial relation, the object categories may cause ambiguity. Thus, we also take the object point clouds as inputs, integrating the geometric features of humans and objects to make the spatial prediction.

\end{itemize}

\subsection*{A.3. Training Details}
\addcontentsline{toc}{subsection}{\textcolor[rgb]{0,0,0}{A.3. Training Details}}

\quad LEMON is implemented by PyTorch and trained with the Adam optimizer. The training epoch is set to $100$. All training processes are on 4 NVIDIA 3090 Ti GPUs with an initial learning rate of $1e$-$4$. The HRNet backbone is initialized with the weights pre-trained on ImageNet \cite{deng2009imagenet}, while the point cloud extractor is trained from scratch. The hyper-parameters $\omega_{1-4}$ that balance loss are set to $40$, $40$, $20$, $20$, respectively, and the training batch size is set to 24.

\section*{B. Dataset}
\label{sub_Dataset}
\addcontentsline{toc}{section}{\textcolor[rgb]{0,0,0}{B. Dataset}}

\quad We curate a question-and-answer (Q \& A) list for readers to have a clearer and more detailed understanding of the 3DIR dataset by referring to Datasheets \cite{gebru2021datasheets}.

\noindent\textbf{Q1: For what purpose was the dataset created? Was there a specific gap that needed to be filled? Please provide a description.}
\par \noindent \textbf{A1:} 3DIR is collected to facilitate research of 3D human-object interaction relation understanding. It contains paired HOI data and annotations of several interaction elements that elucidate ``where'' the interaction manifests between the human and object, \eg, human contact, object affordance, and human-object spatial relation. Most existing datasets enable training task-specific models, making the model perceive a certain element of the human-object interaction in isolation. However, interaction elements manifest in the interacting subject, object, and between them. There are also intricate correlations between the interaction elements. We aspire to endow the model with a more comprehensive capacity to understand the interaction relation between humans and objects. Thus, we collect and annotate the 3DIR. Facilitating the perception of human-object interaction relation by enabling the model to learn joint anticipations of interaction elements.

\noindent\textbf{Q2: How many instances are there in total?}
\par \noindent \textbf{A2:} There are $5001$ in-the-wild images with explicit interaction content, and the quantity of 3D object instances in 3DIR is $5143$. The whole data spans $21$ object classes and $17$ interaction categories. Besides, there are over $25k$ multiple annotations for these collected data.

\noindent\textbf{Q3: What data does each instance consist of?}
\par \noindent \textbf{A3:} A sample of 3DIR contains the following data: 1) an interaction image; 2) an object point cloud; 3) the SMPL-H pseudo-GTs; 4) masks of the interacting human and object in the image; 5) 3D affordance annotation of the object; 6) dense human contact label; 7) spatial position of object centers in the same camera coordinate of the fitted human. In addition, taking the masks could easily calculate the bounding boxes of the human and object.

\begin{figure}
    \centering
    \begin{overpic}[width=1\linewidth]{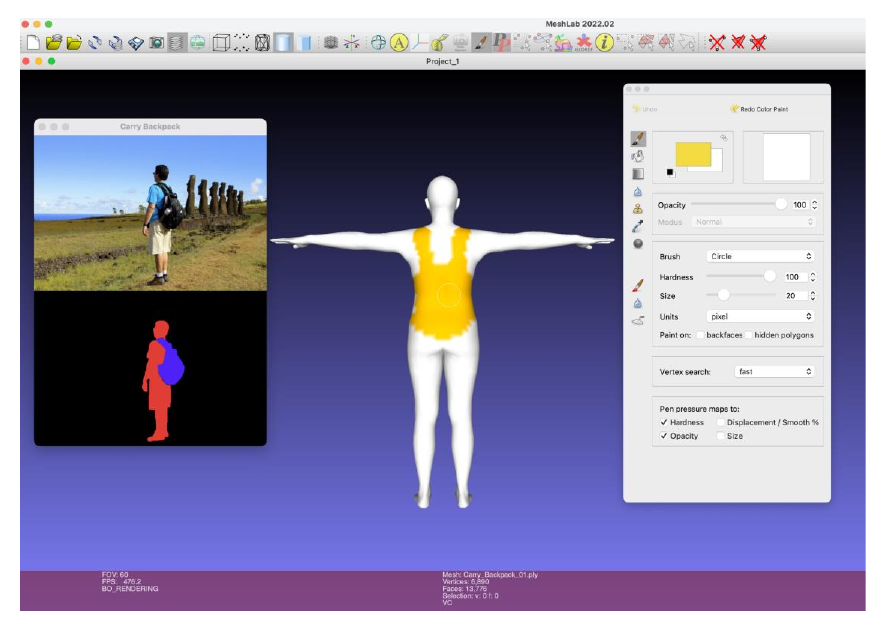}
    
    \end{overpic}
    \caption{Given the image and mask with an interaction type, drawing vertices in contact with the object.}
    \label{fig:sup_contact}
\end{figure}

\noindent\textbf{Q4: Why chose the SMPL-H as the human mesh?}
\par \noindent \textbf{A4:} Firstly, SMPL-H and SMPL \cite{loper2023smpl} share the same topology, so the model trained with SMPL-H could directly generalize to SMPL. However, SMPL lacks parameters for the hands, resulting in some types of grasping interactions that cannot be presented. Another human model SMPL-X \cite{SMPL-X:2019}, also includes hand parameters. But it has $10475$ vertices and nearly $50\%$ of the vertices are on the head, which usually does not greatly impact the interaction. To save the computational overhead, we ultimately chose to use the OSX \cite{lin2023osx} to fit the SMPL-H human mesh.

\noindent\textbf{Q5: Why use the OSX to fit the human body, and how to implement the pipeline?}
\par \noindent \textbf{A5:} In natural interaction images, sometimes only the upper body of a human is visible, and the model should possess the ability to predict the human mesh for these cases. OSX \cite{lin2023osx} is trained on multiple datasets, such as MPII \cite{andriluka14cvpr}, Human3.6M \cite{ionescu2013human3}, AGORA \cite{patel2021agora}. It masters the prior knowledge of human topology with the training on the above datasets. In addition, it provides a pipeline to fit the UBody dataset. The models trained on UBody perform well in predicting situations where only the upper body is present. Therefore, we use the same pipeline as UBody to fit the human mesh of 3DIR. To maximize the utilization of the pre-trained OSX, we still retained the face decoder, and the SMPL-X humans are transferred to SMPL-H through the official script in SMPL-X \cite{SMPL-X:2019}.


\noindent\textbf{Q6: What mechanisms or procedures were used to collect and annotate the data (\eg, software program, software API)?}
\par
\begin{figure}
    \centering
    \begin{overpic}[width=1\linewidth]{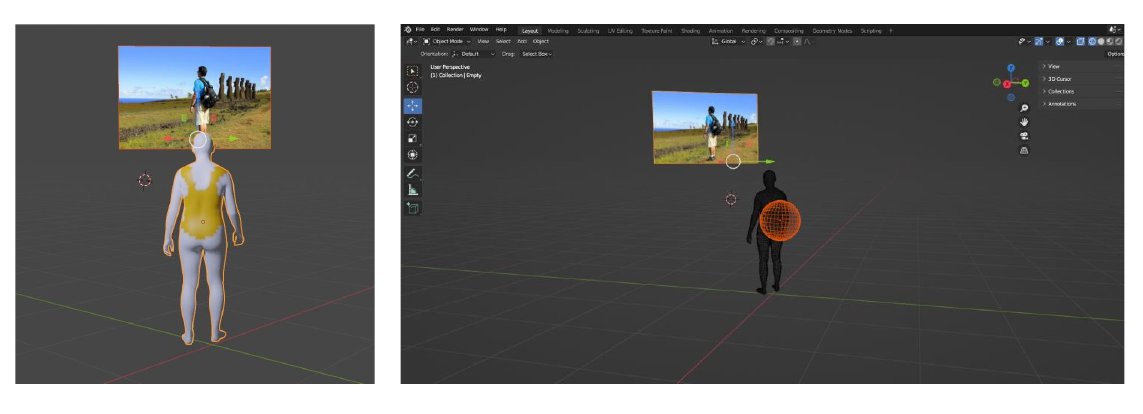}
    
    \end{overpic}
    \caption{Adjust the spatial position of the proxy sphere to align with human-object spatial relation in the image by referring to the image and contact vertices.}
    \label{fig:sup_spatialannotate}
\end{figure}

 \noindent \textbf{A6:} Images are collected from HAKE \cite{li2020pastanet}, V-COCO \cite{gupta2015visual}, PIAD \cite{Yang_2023_ICCV} and websites with free licenses. The 3D object instances mainly come from PartNet \cite{mo2019partnet}, 3D-AffordanceNet \cite{deng20213d} and Objaverse \cite{deitke2023objaverse}. The objects selected from Objaverse are downloaded through its official API. For multiple annotations, we used several tools to implement them respectively. 1) We leverage the \textbf{ISAT} \cite{ISAT} to annotate the human and object masks. It integrates the SAM \cite{kirillov2023segment} and can perform interactive semi-automatic annotation. 2) The software \textbf{MeshLab} \cite{meshlab} is utilized to ``draw'' the dense human contact vertices. In which mesh could be translated, rotated, and zoomed in/out, it also supports drawing colors on the mesh vertices. As shown in Fig. \ref{fig:sup_contact}, we pin the image and mask onto the screen, and annotators color the vertices on the human body that are in contact with the object. The contact vertices are captured based on their color through a script. 3) We refer to the 3D-AffordanceNet \cite{deng20213d} to annotate the object affordance. The object instances are imported to the MeshLab and we color the affordance key points and the propagable region. Their coordinates are recorded, and the remaining propagation steps and algorithms are consistent with 3D-AffordanceNet. Please refer to it for more details. 4) With the contact annotation, we color the fitted human mesh and import it into the \textbf{Blender} \cite{blender} to annotate the object's spatial position. Specifically, we create geometric and material nodes for annotators to view the contact on the human body. In conjunction with the image, annotators need to adjust the position of the sphere (object proxy) with pre-defined radii to align the human-object spatial relation in the image, as shown in Fig. \ref{fig:sup_spatialannotate}. The criteria for defining the radius of the object is clarified in \textbf{Q7}.

\noindent\textbf{Q7: How to define the radius of the object's proxy sphere? Please provide a detailed description.}
\par \noindent \textbf{A7:} We define the object's proxy radius relative to the fitted human mesh. As the fitted human body's unit measurements in the coordinate system match the real world, defining objects' radii in this way closely aligns with the actual size of objects. For each object category, we import $20$ instances into Blender and scale them to match the human size. Then, the center of the proxy sphere is moved to the object's geometric center, adjusting the sphere radius to just envelop the object. We record each radius and calculate the mean as the proxy sphere radius for each object category. Note that these radii could be used as a basis for further tuning through the mask ratio of humans and objects, the ratio of bounding boxes in the image, or others. The defined radii are shown in Tab. \ref{tab:radius}.

\begin{table}[t]
  \centering
  \renewcommand{\arraystretch}{1.}
  \renewcommand{\tabcolsep}{8pt}
    \scriptsize
   \caption{The defined radii for each object category, unit: m.}
   \label{tab:radius}
\begin{tabular}{cc|cc|cc}
\toprule
\textbf{Backpack} & 0.265 & \textbf{Bottle} & 0.140 &  \textbf{Mug} & 0.094\\
\textbf{Baseballbat} & 0.325 & \textbf{Suitcase} & 0.332 &  \textbf{Vase} & 0.197\\
\textbf{Skateboard} & 0.375 & \textbf{Bicycle} & 0.675 &  \textbf{Bowl} & 0.132\\
\textbf{Tennisracket} & 0.298 & \textbf{Scissors} & 0.179 &  \textbf{Chair} & 0.455\\
\textbf{Surfboard} & 0.687 & \textbf{Keyboard} & 0.217 &  \textbf{Knife} & 0.173\\
\textbf{Motorcycle} & 0.710 & \textbf{Earphone} & 0.132 &  \textbf{Bag} & 0.192\\
\textbf{Umbrella} & 0.372 & \textbf{Guitar} & 0.394 &  \textbf{Bed} & 1.154\\
\bottomrule
\end{tabular}
\end{table}

\noindent\textbf{Q8: When fitting the human body, there are already assumed camera optical centers and focal lengths. Why not directly project the center of the object in the 2D image into 3D space?}
\par \noindent \textbf{A8:} The spatial annotation of the object center is conducted within the same camera coordinates as fitted humans. For in-the-wild images, HMR (human mesh recovery) methods mostly take weak-perspective camera models to project the human mesh \cite{horaud1997object, kanazawa2018end} to the image plane. Which considers the depth to be relatively uniform for the human instance. Therefore, directly back-projecting the objects' centers from images into 3D space will cause depth ambiguity. This is also explained in DJ-RN \cite{li2020detailed}, where they further determine the depth of objects by defining their radii. For HOIs, the relative depth between humans and objects is crucial for representing their spatial relation. So, based on the camera scale $s$ and translation $t$ inferred by OSX, we manually annotate the objects' center positions at the same camera coordinates as the fitted humans.

\noindent\textbf{Q9: How to ensure the quality of annotation in the processes?}
\par \noindent \textbf{A9:} We conduct subjective cross-checks and objective measuring to ensure the quality of the annotations. In specific, we initially release the annotation requirements and recruit a cohort of annotators. For the contact annotation, we provide $100$ author-annotated instances, as well as detailed annotation instructions and software tutorials. Then, the annotators make annotations for these $100$ samples following the instructions. Referring to DAMON \cite{tripathi2023deco}, the Intersection-over-Union (IOU) is calculated between the author-annotation and annotations made by candidate annotators. For the objects' spatial positions, we select $100$ instances from the BEHAVE \cite{bhatnagar22behave} dataset and record the coordinates of the objects' geometric centers. The annotators are required to annotate the objects' positions according to the manner in \textbf{Q6}. The MSE is calculated between the ground truth in BEHAVE and annotations. Eventually, we select $5$ qualified annotators through the evaluation results. These $5$ annotators conduct three rounds of annotation, with each round involving subjective cross-check and author-check to filter out instances with glaring annotation errors. Besides, for each instance, we cyclically use the annotation of each annotator as a temporary reference and calculate the measurement metrics (IOU, MSE) between it and the remaining annotations. A re-annotation process will be initiated if there is significant variance among the metrics. We choose the temporary annotation with the minimum variance in metrics as the final annotation. The object affordances are annotated by authors. We train IAG-Net \cite{Yang_2023_ICCV} on $10$ object categories we annotated and $11$ categories selected from 3D-AffordanceNet. The AUC and aIOU for our annotated data are $85.15, 37.93$, while for the selected data are $85.76, 37.82$. This indirectly indicates that the quality of our annotations is comparable to existing annotations, which could effectively support the model training.

\noindent\textbf{Q10: Are there any errors, noise, or redundancies in the dataset? If so, please provide a description.}
\par \noindent \textbf{A10:} Since some annotations are based on human knowledge, \eg, human contact, some annotations may not be completely accurate. However, these will not seriously impact the final results. Similar to those instances in the DAMON \cite{tripathi2023deco} and HOT \cite{chen2023hot} dataset.

\noindent\textbf{Q11: Does the dataset contain data that might be considered confidential (e.g., data that is protected by legal privilege or includes the content of individuals non-public communications)?}
\par \noindent \textbf{A11:} No.

\noindent\textbf{Q12: Does the dataset contain data that, if viewed directly, might be offensive, insulting, threatening, or might otherwise cause anxiety?}
\par \noindent \textbf{A12:} No.

\section*{C. Experiments}
\addcontentsline{toc}{section}{\textcolor[rgb]{0,0,0}{C. Experiments}}

\quad We conduct more experiments to verify the superiority of the LEMON. The details are described as follows.

\subsection*{C.1. Test on Multiple Datasets}
\addcontentsline{toc}{subsection}{\textcolor[rgb]{0,0,0}{C.1. Test on Multiple Datasets}}

\quad DAMON Dataset \cite{tripathi2023deco}. We select around $3k$ data from DAMON that match the object categories in 3DIR, divide the training and testing sets in an $8:2$ ratio and train the comparison methods. The training of DECO here is consistent with its original architecture, using the scene, part, and contact branches. The human meshes used in LEMON are directly inferred through OSX accompanied by a transferring process, without the fitting pipeline that is illustrated in Sec. B \textbf{Q5}. Evaluation results are recorded in Tab. \ref{tab:damon}.

\begin{table}[t]
  \centering
  \renewcommand{\arraystretch}{1.}
  \renewcommand{\tabcolsep}{8pt}
   \caption{Comparison of contact estimation on DAMON dataset.}
   \label{tab:damon}
\begin{tabular}{c|cccc}
\toprule
\textbf{Method} & \textbf{Precision} & \textbf{Recall} & \textbf{F1} &  \textbf{geo. (cm)}\\ \midrule
BSTRO & 0.57 & 0.58 & 0.53 & 28.81\\
DECO & 0.64 & 0.60 & 0.59 & 19.98\\
Ours P. & 0.69 & 0.62 & 0.61 & 14.75\\
Ours D. & 0.67 & 0.66 & 0.63 & 13.45\\
\bottomrule
\end{tabular}
\end{table}

\begin{table}[t]
\footnotesize
  \renewcommand{\arraystretch}{1.}
  \renewcommand{\tabcolsep}{4.pt}
  \caption{{Take BEHAVE as the unseen dataset. \textbf{(a)} Comparison of contact estimation on BEHAVE dataset. \textbf{(b)} Comparison of spatial prediction on BEHAVE dataset.}}
\label{table:behave}
\begin{subtable}[t]{0.72\linewidth}
\begin{tabular}{c|cccc}
\toprule
\textbf{Method} & \textbf{Precision} & \textbf{Recall} & \textbf{F1} &  \textbf{geo. (cm)}\\ \midrule
BSTRO & 0.03 & 0.11 & 0.05 & 48.37\\
DECO & 0.13 & 0.25 & 0.17 & 40.45\\
Ours P. & 0.17 & 0.27 & 0.20 & 35.14\\
Ours D. & 0.19 & 0.30 & 0.21 & 29.82\\
\bottomrule
\end{tabular}
\caption{}
\end{subtable}
\begin{subtable}[t]{0.25\linewidth}
\centering
\begin{tabular}{c|c}
\toprule
\textbf{Method} & \textbf{MSE} \\ \midrule
DJ-RN & 0.287 \\
PopUp & 0.149 \\
Ours P. & 0.098 \\
Ours D. & 0.084 \\
\bottomrule
\end{tabular}
\caption{}
\end{subtable}
\vspace{-10pt}
\end{table}

\begin{table}[t]
\footnotesize
  \centering
  \renewcommand{\arraystretch}{1.}
  \renewcommand{\tabcolsep}{7.pt}
   \caption{Object affordance anticipation on PIAD dataset. Seen and Unseen are two settings, 3D-Aff. denotes 3D-AffordanceNet.}
   \label{tab:PIAD}
\begin{tabular}{c|ccc|ccc}
\toprule
\multirow{2}{*}{\textbf{Method}} & \multicolumn{3}{c|}{\textbf{Seen}}  & \multicolumn{3}{c}{\textbf{Unseen}}\\ \cmidrule{2-7}
 & \textbf{AUC} & \textbf{aIOU} & \textbf{SIM} & \textbf{AUC} & \textbf{aIOU} & \textbf{SIM} \\ \midrule
3D-Aff. & 83.07 & 16.65 & 0.46 & 59.51 & 3.87 & 0.323\\
IAG & 84.85 & 20.51 & 0.54 & 64.14& 7.35 & 0.346\\
Ours P. & 85.64 & 22.98 & 0.56 & 65.96& 7.88 & 0.351\\
Ours D. & 86.07 & 23.16 & 0.56 & 66.22& 8.23 & 0.355\\
\bottomrule
\end{tabular}
\end{table}

BEHAVE Dataset \cite{bhatnagar22behave}. The BEHAVE dataset is utilized as an unseen dataset to evaluate both the contact estimation and spatial prediction. For the object position, we record the geometry center of aligned objects in the BEHAVE. Each comparison method is only trained on the 3DIR dataset and tested on the BEHAVE dataset. The metrics of both the contact and spatial prediction are shown in Tab. \ref{table:behave}.

PIAD Dataset \cite{Yang_2023_ICCV}. We conduct tests with two settings on PIAD. 1) Seen: as many images in the PIAD dataset do not include complete humans, the template human is utilized to train LEMON on PIAD. 2) Unseen: PIAD and 3DIR have $11$ overlapping object categories, so we train methods on $10$ categories in 3DIR (not included in PIAD) and test them on the $11$ categories (regarded as unseen data) in PIAD. The results of ``Seen'' and ``Unseen'' settings are recorded in Tab. \ref{tab:PIAD}.

Following DECO \cite{tripathi2023deco} and HOT \cite{chen2023hot}, we evaluate whether the estimated contact by LMEON benefits human pose and shape (HPS) regression. The test is on the PROX “quantitative” dataset \cite{hassan2019resolving}, and the experimental setup is the same with DECO and HOT. Since LEMON focuses on estimating vertices that are in contact with objects, the neglect of estimating the contact between feet and the ground has a certain impact on the results. Thus, we fine-tune LEMON on the DAMON dataset and give the results in Tab. \ref{tab:prox}.

\begin{table}[]
\footnotesize
\centering
  \renewcommand{\arraystretch}{1.}
  \renewcommand{\tabcolsep}{4.pt}
   \caption{HPS estimation performance using contact derived from different sources.}
    \label{tab:prox}
\begin{tabular}{c|cccccc}
\toprule
  \textbf{Methods}  & \ding{55} \textbf{Contact} & \textbf{Prox} & \textbf{HOT} & \textbf{DECO} & \textbf{LEMON} & \textbf{GT}\\ \midrule
\textbf{V2V $\downarrow$} & 183.3 & 174.0 & 172.3 & 171.6 & \textbf{170.9} & 163.0 \\
\bottomrule
\end{tabular}
\end{table}

\begin{table}[]
\footnotesize
  \centering
  \renewcommand{\arraystretch}{1.}
  \renewcommand{\tabcolsep}{4.pt}
   \caption{The influence of learning rate (Lr) and batch size (B). The best results are covered with the mask.}
   \label{tab:Lr}
\begin{tabular}{cc|cccc|ccc|c}
\toprule
\multirow{2}{*}{\textbf{Lr}} & \multirow{2}{*}{\textbf{B}} & \multicolumn{4}{c|}{\textbf{Contact}}                    & \multicolumn{3}{c|}{\textbf{Affordance}}    & \textbf{Spatial} \\ \cmidrule{3-10} 
                             &                             & \textbf{Pre.} & \textbf{Rec.} & \textbf{F1} & \textbf{geo.} & \textbf{AUC} & \textbf{aIOU} & \textbf{SIM} & \textbf{MSE}     \\ \midrule
1e-3        &              16               &        0.76      &       0.77       &       0.75      &       8.67       &       87.26        &      40.52          &       0.59       &         0.017         \\
1e-4                             &            16                 &       0.76       &     0.80         &       0.76      &       7.98       &       87.97       &       40.86        &      0.62        &       0.013           \\
1e-5 & 16 &   0.78 &      0.79        &     0.76        &      7.85        &      88.02        &        41.10       &      0.62        &       0.014           \\
1e-3 &  24  &        0.77      &       0.77       &       0.78      &       7.86       &      87.69        &      40.77          &       0.62       &         0.013         \\
\cellcolor{mygray}1e-4        &              \cellcolor{mygray}24               &        \cellcolor{mygray}0.78      &       \cellcolor{mygray}0.82       &       \cellcolor{mygray}0.78      &       \cellcolor{mygray}7.55       &       \cellcolor{mygray}88.51        &      \cellcolor{mygray}41.34          &       \cellcolor{mygray}0.64       &         \cellcolor{mygray}0.010         \\
1e-5        &              24               &        0.76      &       0.75       &       0.75      &       8.14       &       87.57        &      40.63          &       0.59       &         0.016         \\
1e-3        &              32               &        0.79      &       0.80       &       0.78      &       7.63       &       88.35        &      41.28          &       0.64       &         0.010         \\
1e-4        &              32               &        0.78      &       0.80       &       0.77      &       7.67       &       88.29        &      41,02          &       0.62       &         0.012         \\
\bottomrule
\end{tabular}
\end{table}

\begin{table}[]
\footnotesize
  \centering
  \renewcommand{\arraystretch}{1.}
  \renewcommand{\tabcolsep}{4.pt}
   \caption{Performance of the model under different quantities of human vertices.}
   \label{tab:vertices}
\begin{tabular}{c|cccc|ccc|c}
\toprule
\multirow{2}{*}{\textbf{Vertices}} & \multicolumn{4}{c|}{\textbf{Contact}}                    & \multicolumn{3}{c|}{\textbf{Affordance}}    & \textbf{Spatial} \\ \cmidrule{2-9} 
                             & \textbf{Pre.} & \textbf{Rec.} & \textbf{F1} & \textbf{geo.} & \textbf{AUC} & \textbf{aIOU} & \textbf{SIM} & \textbf{MSE}     \\ \midrule
431      &        0.75      &       0.78       &       0.74      &       12.13       &       87.52        &      39.89          &       0.61       &         0.012         \\
1723                 &       0.78       &     0.82         &       0.78      &       7.55       &       88.51       &       41.34        &      0.64        &       0.010           \\
6890 &   0.79 &      0.80        &     0.78        &      7.62        &      88.32        &        41.53       &      0.63        &       0.009           \\
\bottomrule
\end{tabular}
\end{table}

\begin{table}[]
\footnotesize
\centering
  \renewcommand{\arraystretch}{1.}
  \renewcommand{\tabcolsep}{6.pt}
   \caption{Performance of object spatial position prediction when using different positional encoding methods. P.E. indicates positional encoding.}
    \label{tab:pe}
\begin{tabular}{c|cccc}
\toprule
    & $\bm{w/o}$ \textbf{P.E.} & \textbf{Learnable} & \textbf{Sine \& Cosine} & \textbf{Relative} \\ \midrule
\textbf{MSE} & 0.017 & 0.010 & 0.012 & 0.014 \\
\bottomrule
\end{tabular}
\vspace{-10pt}
\end{table}

\subsection*{C.2. Hyperparameters}
\addcontentsline{toc}{subsection}{\textcolor[rgb]{0,0,0}{C.2. Hyperparameters}}

\begin{table*}[]
\footnotesize
\centering
  \renewcommand{\arraystretch}{1.}
  \renewcommand{\tabcolsep}{1.5pt}
   \caption{Evaluation metrics for each object. Ear. is Earphone, Baseb. is Baseballbat, Tenn. is Tennisracket, Motor. is Motorcycle, Back. is Backpack, Kni.is Knife, Bicy. is Bicycle, Umbr. is Umbrella, Keyb. is Keyboard, Bott. is Bottle, Surf. is Surfboard, Suitc. is Suitcase, Skate. is Skateboard. P. indicates take PointNet++ as the point cloud backbone while D. indicates DGCNN.}
    \label{tab:per_object}
\begin{tabular}{c|c|ccccccccccccccccccccc}
\toprule
                 & \textbf{Metr.} & \textbf{Ear.} & \textbf{Baseb.} & \textbf{Tenn.} & \textbf{Bag} & \textbf{Motor.} & \textbf{Gui.} & \textbf{Back.} & \textbf{Chair} & \textbf{Kni.} & \textbf{Bicy.} & \textbf{Umbr.} & \textbf{Keyb.} & \textbf{Scis.} & \textbf{Bott.} & \textbf{Bowl} & \textbf{Surf.} & \textbf{Mug} & \textbf{Suitc.} & \textbf{Vase} & \textbf{Skate.} & \textbf{Bed} \\ \midrule
\multirow{8}{*}{\rotatebox[origin=c]{90}{\textbf{LEMON P.}}} & \textbf{Prec.}   & 0.97          & 0.75              & 0.80           & 0.29         & 0.90            & 0.73            & 0.72           & 0.77           & 0.75           & 0.84           & 0.70           & 0.67            & 0.35           & 0.71           & 0.76          & 0.85           & 0.67         & 0.93            & 0.73          & 0.79            & 0.75         \\
                                   & \textbf{Rec.}    & 0.94          & 0.87              & 0.86           & 0.43         & 0.88            & 0.75            & 0.69           & 0.89           & 0.84           & 0.86           & 0.86           & 0.84            & 0.54           & 0.83           & 0.81          & 0.84           & 0.59         & 0.94            & 0.84          & 0.94            & 0.77         \\
                                   & \textbf{F1}      & 0.95          & 0.79              & 0.8            & 0.33         & 0.88            & 0.72            & 0.69           & 0.79           & 0.78           & 0.84           & 0.76           & 0.73            & 0.34           & 0.73           & 0.75          & 0.84           & 0.60         & 0.94            & 0.76          & 0.84            & 0.74         \\
                                   & \textbf{geo.}     & 1.54          & 2.94              & 13.75          & 37.27        & 2.34            & 3.30            & 7.25           & 3.97           & 16.39          & 6.26           & 22.08          & 18.65           & 32.75          & 24.19          & 3.37          & 5.70            & 19.77        & 0.25            & 3.39          & 3.91            & 4.38         \\ \cmidrule{2-23} 
                                   & \textbf{AUC}     & 86.85         & 94.41             & 97.09          & 92.08        & 97.67           & 96.28           & 90.49          & 94.32          & 82.67          & 83.78          & 95.78          & 86.2            & 64.86          & 68.44          & 68.55         & 77.43          & 82.75        & 91.4            & 69.77         & 91.98           & 89.71        \\
                                   & \textbf{aIOU}    & 23.71         & 63.12             & 58.12          & 38.76        & 51.64           & 71.05           & 50.8           & 36.18          & 9.62           & 35.50          & 59.24          & 15.06           & 5.12           & 11.99          & 3.63          & 42.05          & 32.029       & 50.42           & 5.02          & 76.21           & 23.40        \\
                                   & \textbf{SIM}     & 0.61          & 0.74              & 0.70           & 0.47         & 0.63            & 0.82            & 0.64           & 0.71           & 0.60           & 0.44           & 0.66           & 0.26            & 0.41           & 0.60           & 0.79          & 0.39           & 0.64         & 0.55            & 0.67          & 0.85            & 0.59         \\ \cmidrule{2-23} 
                                   & \textbf{MSE}     & 0.016         & 0.012             & 0.028          & 0.022        & 0.012           & 0.009           & 0.032          & 0.01           & 0.008          & 0.007          & 0.010          & 0.008           & 0.008          & 0.006          & 0.004         & 0.012          & 0.008        & 0.018           & 0.003         & 0.008           & 0.048        \\ \midrule
\multirow{8}{*}{\rotatebox[origin=c]{90}{\textbf{LEMON D.}}} & \textbf{Prec.}   & 0.95          & 0.77              & 0.83           & 0.38         & 0.89            & 0.74            & 0.77           & 0.81           & 0.80           & 0.82           & 0.74           & 0.69            & 0.37           & 0.75           & 0.80          & 0.85           & 0.68         & 0.95            & 0.74          & 0.80            & 0.78         \\
                                   & \textbf{Rec.}    & 0.95          & 0.87              & 0.84           & 0.46         & 0.89            & 0.73            & 0.69           & 0.86           & 0.79           & 0.89           & 0.87           & 0.79            & 0.67           & 0.78           & 0.81          & 0.82           & 0.76         & 0.94            & 0.82          & 0.89            & 0.79         \\
                                   & \textbf{F1}      & 0.95          & 0.80              & 0.82           & 0.40         & 0.88            & 0.71            & 0.69           & 0.80           & 0.78           & 0.84           & 0.78           & 0.71            & 0.42           & 0.74           & 0.78          & 0.82           & 0.69         & 0.94            & 0.76          & 0.83            & 0.77         \\
                                   & \textbf{geo.}     & 0.70          & 2.47              & 12.18          & 31.53        & 2.78            & 3.51            & 4.02           & 3.11           & 9.94           & 6.62           & 23.97          & 16.09           & 27.43          & 17.91          & 3.59          & 5.86           & 11.87        & 0.04            & 2.92          & 1.95            & 2.73         \\ \cmidrule{2-23} 
                                   & \textbf{AUC}     & 87.90         & 97.45             & 98.99          & 93.04        & 97.92           & 97.74           & 95.49          & 94.23          & 82.02          & 87.10          & 96.88          & 86.67           & 75.96          & 69.16          & 66.69         & 73.67          & 83.16        & 84.83           & 72.08         & 93.64           & 87.66        \\
                                   & \textbf{aIOU}    & 22.01         & 69.37             & 71.03          & 43.84        & 53.04           & 72.65           & 60.33          & 36.68          & 9.56           & 31.09          & 60.59          & 16.72           & 5.46           & 12.76          & 3.48          & 41.36          & 32.94        & 41.45           & 5.29          & 79.70           & 21.63        \\
                                   & \textbf{SIM}     & 0.61          & 0.81              & 0.78           & 0.48         & 0.64            & 0.84            & 0.7            & 0.69           & 0.61           & 0.40           & 0.69           & 0.28            & 0.46           & 0.60           & 0.79          & 0.37           & 0.65         & 0.51            & 0.68          & 0.87            & 0.57         \\ \cmidrule{2-23} 
                                   & \textbf{MSE}     & 0.002         & 0.007             & 0.025          & 0.019        & 0.008           & 0.007           & 0.021          & 0.011          & 0.005          & 0.007          & 0.012          & 0.006           & 0.004          & 0.006          & 0.003         & 0.008          & 0.003        & 0.030           & 0.002         & 0.005           & 0.040        \\ \bottomrule
\end{tabular}
\end{table*}

\quad During the training process, some hyperparameters have impacts on the model performance. We provide a series of experiments to determine the ultimate hyperparameters. All experiments are conducted when taking DGCNN as the point cloud backbone. Concerning the impact of learning rate and batch size on the model, we conduct comparative experiments by adjusting the learning rate across orders of magnitude and combining it with different batch sizes. The results are presented in Tab. \ref{tab:Lr}. Besides, we test whether the quantity of human vertices influences the model performance. The quantity of object points is consistent with 3D-AffordanceNet and IAG-Net. The results are shown in Tab. \ref{tab:vertices}. As can be seen, when the number of vertices increases from $431$ to $1723$, there is a significant increase in model performance. However, the growth is not significant when increasing from $1723$ to $6890$. To conserve computational overhead, we ultimately chose to sample the number of human vertices at $1723$. We also test the performance of several position encoding methods in object spatial position prediction. The results are reported in Tab. \ref{tab:pe}. For $w_{1-4}$ that balance the loss, we test them according to the order of magnitude and multiples, and finally determine their specific values. Due to the excessive number of combinations, we do not exhibit the results one by one here.


\subsection*{C.3. More Results}
\addcontentsline{toc}{subsection}{\textcolor[rgb]{0,0,0}{C.3. More Results}}

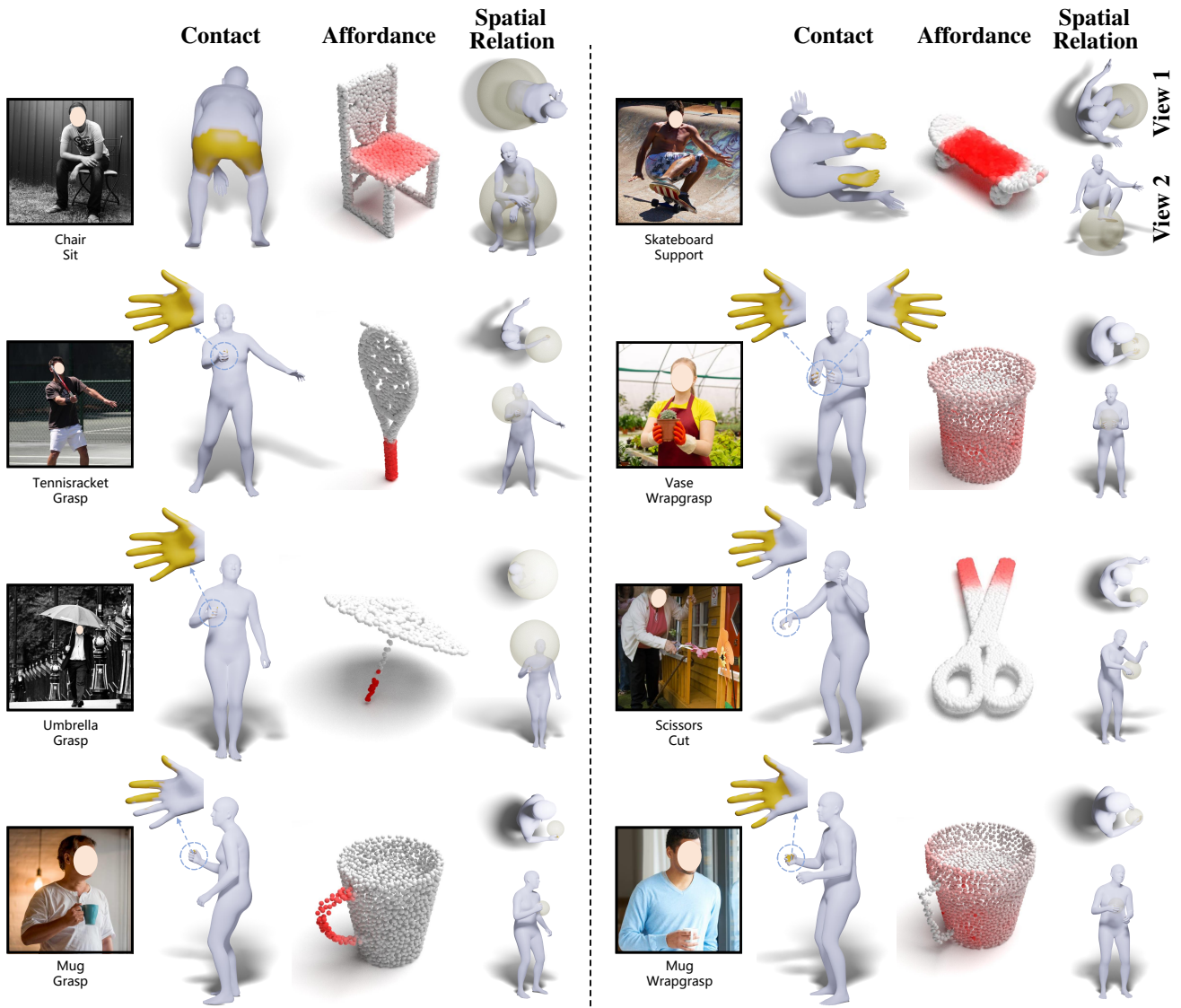
\begin{figure*}[t]
    \centering
    \begin{overpic}[width=1\linewidth]{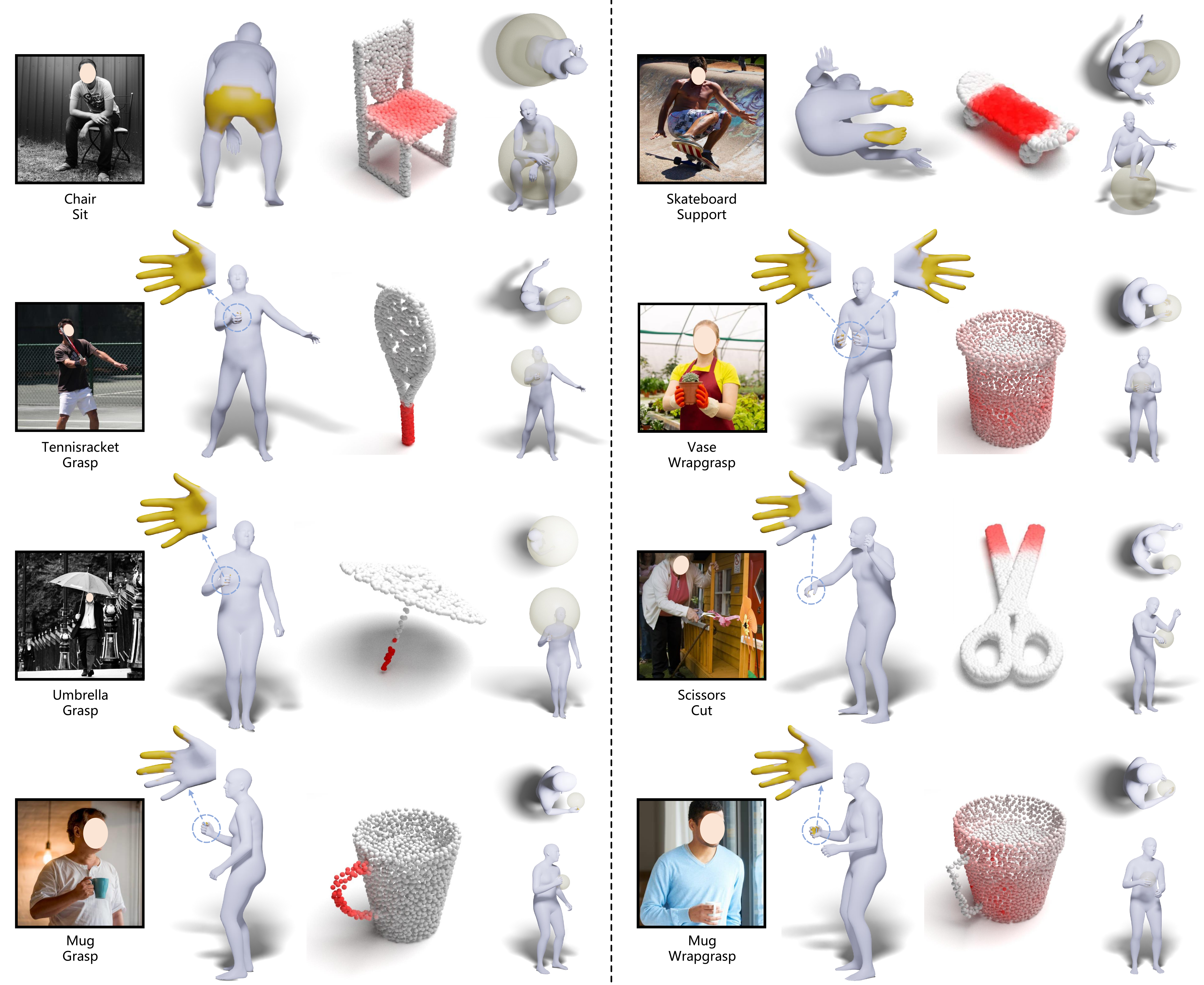}
    \put(16,82){\textbf{Contact}}
    \put(28,82){\textbf{Affordance}}
    \put(41,83.5){\textbf{Spatial}}
    \put(40.5,81.5){\textbf{Relation}}

    \put(68,82){\textbf{Contact}}
    \put(78.5,82){\textbf{Affordance}}
    \put(90.5,83.5){\textbf{Spatial}}
    \put(90,81.5){\textbf{Relation}}

    \put(98.5,74){\rotatebox{90}{\textbf{View 1}}}
    \put(98.5,65){\rotatebox{90}{\textbf{View 2}}}
    
    \end{overpic}
    \caption{More visual results anticipated by LEMON, including human contact, object affordance, and human-object spatial relation.}
    \label{fig:sup_result1}
\end{figure*}

\begin{figure*}[t]
    \centering
    \begin{overpic}[width=0.9\linewidth]{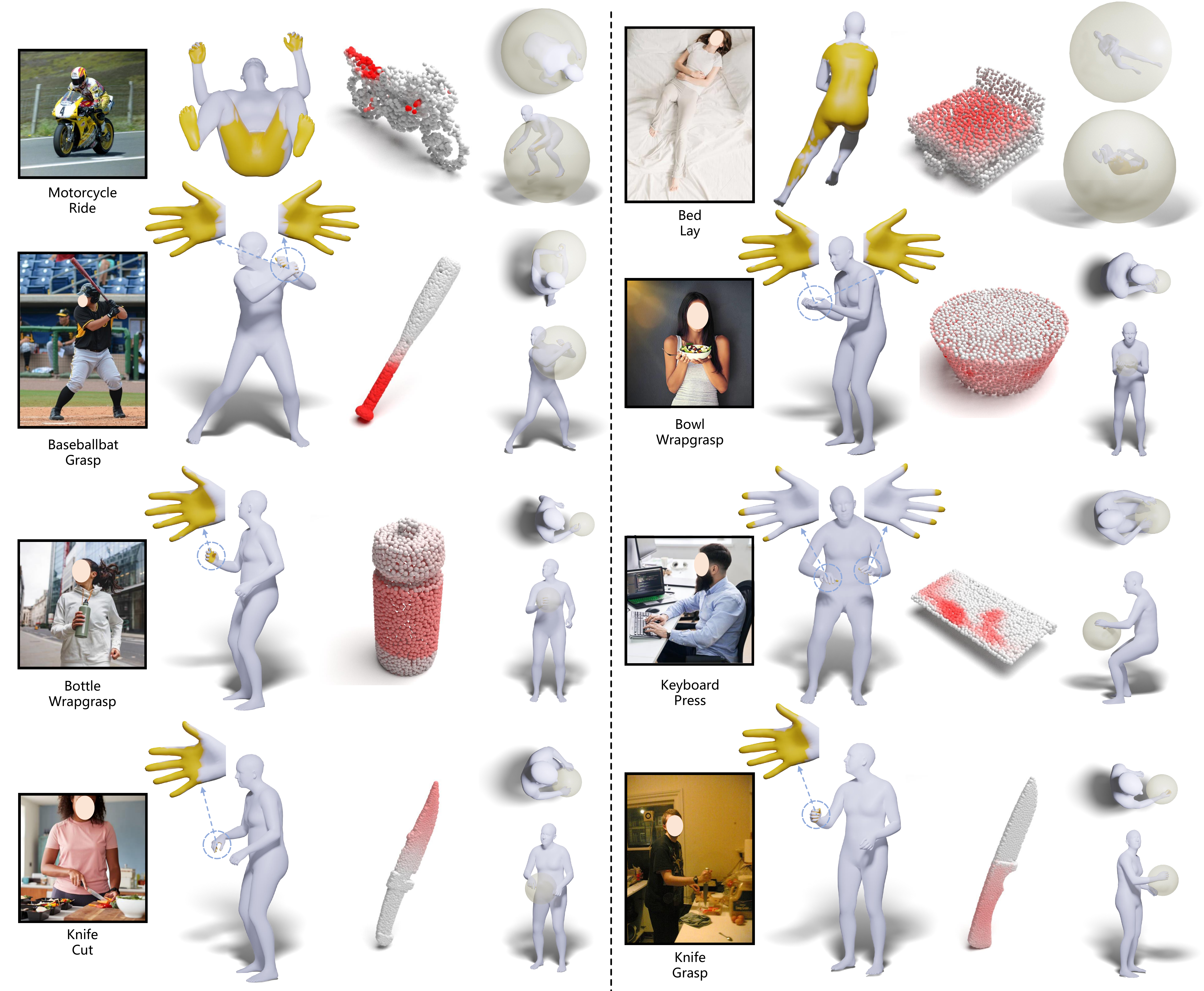}
    \put(17,83){\textbf{Contact}}
    \put(28,83){\textbf{Affordance}}
    \put(41.5,84.5){\textbf{Spatial}}
    \put(41,82.5){\textbf{Relation}}

    \put(66,83){\textbf{Contact}}
    \put(76,83){\textbf{Affordance}}
    \put(89.5,84.5){\textbf{Spatial}}
    \put(89,82.5){\textbf{Relation}}

    \put(99,75){\rotatebox{90}{\textbf{View 1}}}
    \put(99,66){\rotatebox{90}{\textbf{View 2}}}
    
    \end{overpic}
    \caption{More visual results anticipated by LEMON, including human contact, object affordance, and human-object spatial relation.}
    \label{fig:sup_result2}
\end{figure*}

\begin{figure*}[t]
    \centering
    \begin{overpic}[width=0.9\linewidth]{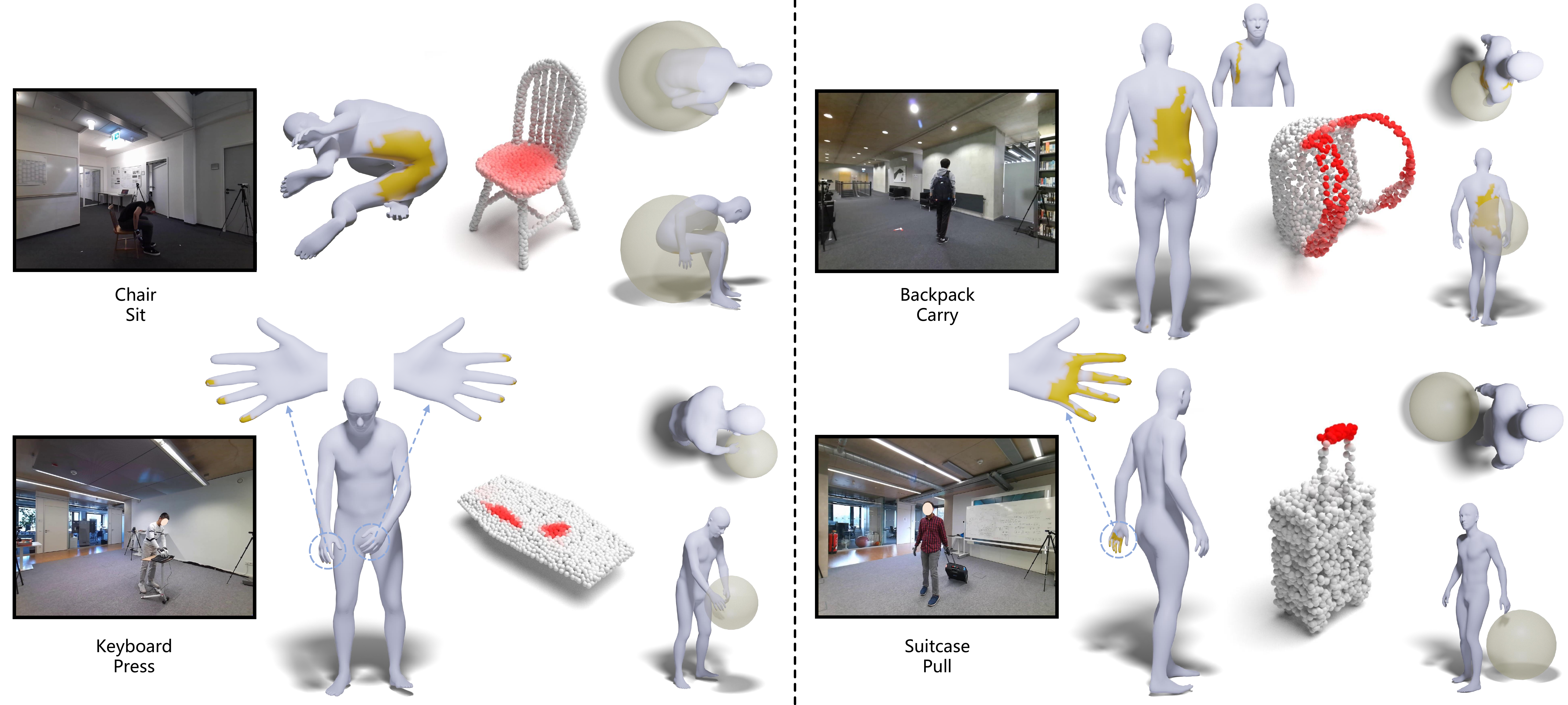}
    
    \end{overpic}
    \caption{Anticipating on the unseen BEHAVE data, including human contact, object affordance, and human-object spatial relation.}
    \label{fig:sup_behave}
\end{figure*}

\quad In the main paper, we provide overall results of metrics and some visualization results of LEMON on the 3DIR benchmark. Here, we show evaluation metrics for each object category and more visualization results that are not presented in the main paper. The metrics for each object are shown in Tab. \ref{tab:per_object}. Fig. \ref{fig:sup_result1} and Fig. \ref{fig:sup_result2} demonstrate more visual results, including human contact, object affordance, and spatial relation. For the experiments in Tab. \ref{table:behave}, we also provide some visual results of LEMON, shown in Fig. \ref{fig:sup_behave}.


\section*{D. Application Prospect}
\addcontentsline{toc}{section}{\textcolor[rgb]{0,0,0}{D. Application Prospect}}

\quad The human contact, object affordance, and human-object spatial relation are crucial elements for representing the human-object interaction relation. Perceiving these elements also links the HOI understanding with downstream applications.

\textbf{Embodied AI} \cite{savva2019habitat}. One characteristic of embodied intelligence is to learn and improve skills by actively interacting with the surrounding environment \cite{nagarajan2020learning}. However, the prerequisite for actively interacting with the environment is the ability to perceive or understand how to interact \cite{fang2018demo2vec, nagarajan2019grounded}. Learning and understanding interactions from human-object interaction is an effective manner. The interaction elements reflect how the interaction is manifest at the counterparts. For example, object affordance represents what action could be done for the object and which location supports the action, revealing ``where to interact''. Human contact represents the regions capable of interacting with objects on the embodiment, revealing ``where are utilized to interact''. Spatial relation connects the interacting subject and object. These elements collectively formulate the interaction relation. Perceiving interaction elements enables the embodied agent to make policies on how to interact with the environment, thereby learning from interactions.

\textbf{Interaction Modeling}. Modeling or recovering the interaction is a significant but extremely challenging task. It holds significant prospects for application in the animation and digital avatar industries. Some methods have explored the benefits of human contact in interaction modeling \cite{xie2022chore, xu2021d3d, hassan2021populating, zhang2020perceiving, xu2023interdiff}. But obviously, human contact is only one aspect of the interaction relation. In addition to this, object affordance and human-object spatial relation also offer clues for interaction modeling. The exploration of incorporating more comprehensive representations of interaction relation (\eg, affordance, spatial relation) into interaction modeling is a worthy investigated future research direction. Which may further advance the modeling of interactions.

\textbf{Augement \& Virtual Reality}. With the emergence of immersive spatial computing devices, \eg, Meta Quest, Apple Vision Pro \cite{waisberg2023future}, and PICO, AR/VR will permeate many industries such as education, healthcare, gaming, and so on. The way of human-computer interaction (HCI) \cite{carroll1997human, mackenzie2012human} is a crucial symptom node for these devices. Perceiving 3D human-object interaction relation in the virtual or augmented world provides feedback signals to adjust the manner of HCI, thereby enhancing the user's immersion.

\textbf{Imitation Learning} \cite{hussein2017imitation}. Imitation learning is an important way to drive robots to complete certain tasks, which makes intelligent agents perform interactions by observing demonstrations from humans or other sources. The interaction elements perceived from 3D human-object interactions offer explicit representations to reveal ``what'' interaction could be performed with an object and ``how'' to interact with it. These rich interaction priors enhance the machine's ability to imitate the interaction manners, thereby learning skills from them. Which is beneficial for configurations like dexterous hand \cite{andrychowicz2020learning, liu2008multisensory} and humanoid robot \cite{kaneko2008humanoid, nelson2018petman, breazeal2003emotion}.

\clearpage
{
    \small
    \bibliographystyle{ieeenat_fullname}
    \bibliography{main}
}

\end{document}